\documentclass{article}

\PassOptionsToPackage{numbers}{natbib}

\usepackage[preprint]{neurips_2026}

\usepackage[utf8]{inputenc} %
\usepackage[T1]{fontenc}    %
\usepackage{url}            %
\usepackage{booktabs}       %
\usepackage{amsfonts}       %
\usepackage{nicefrac}       %
\usepackage{microtype}      %
\usepackage{xcolor}         %
\usepackage{graphicx}
\usepackage{amsmath}
\usepackage{hyperref}
\usepackage[capitalise]{cleveref}
\usepackage{hyperref}
\usepackage{xcolor}
\hypersetup{colorlinks=true, allcolors=blue}
\usepackage{caption}
\usepackage{multirow}
\usepackage{comment}
\usepackage{longtable}

\makeatletter
\AddToHook{cmd/appendix/before}{\def\cref@section@alias{appendix}\def\cref@subsection@alias{appendix}}
\makeatother

\newif\iffinal
\finalfalse   %

\iffinal
    \newcommand{\fazl}[1]{}
\else
    \newcommand{\fazl}[1]{\textcolor{blue}{Fazl: #1}}
\fi

\title{Pretraining Curricula Enable Selective Fine-tuning}

\author{%
  Sebastian A.~Bruijns\thanks{Equal authorship, randomized order. Correspondence to: jirko.rubruck@stx.ox.ac.uk, mia.whitefield@linacre.ox.ac.uk, kai.sandbrink@psy.ox.ac.uk, sebastian.bruijns@psy.ox.ac.uk} \\
  University of Oxford\\
  \And
  Jirko Rubruck$^*$\\
  University of Oxford \\
  \And
  Mia H.~Whitefield$^*$\\
  University of Oxford \\
  \And
  Kai J.~Sandbrink$^*$ \\
  University of Oxford \\
  \And
  Fazl Barez \\
  University of Oxford \& Martian
  \And
  Christopher Summerfield \\
  University of Oxford \\
}

\begin{document}

\maketitle

\begin{abstract}

Transformers follow \emph{implicit} curricula whereby some tasks are learned before others. However, how \emph{explicit} pretraining curricula influence learning, generalization, and the selectivity of fine-tuning is unclear. This is important for AI safety, where fine-tuning is used to selectively suppress misaligned behaviors. Here, we compare curricula that pretrain tasks in a balanced (sampled uniformly) or an imbalanced (one task early, the other late) fashion. We show that imbalanced learning of two conflicting copy tasks promotes in-context learning and improves the selectivity of refusal fine-tuning. Ablations and activation patching show that this occurs because imbalanced pretraining encourages tasks to be disentangled in separable neural circuits, whereas balanced training routes both tasks through a common pathway. We extend these findings to a synthetic language learning task involving rule-consistent and rule-violating data, where imbalanced curricula similarly lead to more localized, less entangled rule representations, resulting in more robust rule-following behavior. Together, these results suggest that imbalanced pretraining curricula may be an important tool for promoting disentangled representations, with direct consequences for the precision and reliability of safety fine-tuning.

\end{abstract}

\section{Introduction}

Neural networks trained with gradient descent learn representations in a progressive, stage-like manner \citep{saxe2019mathematical, zhang2025SaddletoSaddleDynamicsExplains}. Theoretical work has shown that neural networks learn linear before nonlinear functions \citep{kalimeris_sgd_2019}, lower before higher frequencies \citep{rahaman_spectral_2019}, and major before minor modes of the input-output correlation matrix \citep{saxe2019mathematical, rubruck_early_2025}. Recently, these observations have been extended to large language models (LLMs), which learn unigrams before bigrams \citep{belrose2024NeuralNetworksLearn} and simple string operations before logic and complex reasoning \citep{liu2026WhatLanguageModels, michaud2023QuantizationModelNeural}. Interestingly, when examining the progression of learning in LLMs, skills that emerge at similar points in training are co-localised in the residual stream \citep{liu2026WhatLanguageModels}, suggesting that learning order may shape not only \emph{when} capabilities emerge, but also \textit{how} they are internally organized~\citep{geva_transformer_2021}. Thus, like in other deep networks, learning in LLMs is highly path dependent, with acquisition of advanced abilities frequently being layered on top of prerequisite skills \citep{chen2024SkillsinContextUnlockingCompositionality, chen2023SkillitDataDrivenSkills, lee2025DistinctComputationsEmerge}.

Despite this evidence that deep networks follow an \emph{implicit} curriculum for representation learning, attempts to promote learning in deep networks by imposing an \emph{explicit} curriculum have been mixed. Structure in training examples has been argued to improve robustness to noise \citep{volk2025CurriculumEffectVisual} and to boost pretraining efficiency in LLMs \citep{zhang2026RandomSamplingEfficient}, and curricula help bridge the gap between actions and outcomes in reinforcement learning (RL) \citep{lee2024WhyAnimalsNeed,krueger2009FlexibleShapingHow}. At the same time, the literature is dogged by numerous failures \citep{wu_when_2021}. This contrasts sharply with investigations of biological brains, which follow implicit curricula \citep{mandler_concept_1993} and near-universally benefit from explicit curricula, especially those which proceed from generalities to specifics \citep{mathy_rule-based_2009} or present information that is correlated over time \citep{flesch_orthogonal_2022, noh2016OptimalSequencingCategory}. Despite emerging evidence that in-context learning (ICL) in transformer-based networks may display some of the curriculum dependence observed in humans \citep{russin2025Parallel_trade-offs} we still lack understanding of when curricula may work in deep networks \citep{mannelli_tilting_2024}.

Here, we ask how pretraining curricula shape representation learning in transformers, with a focus on entanglement of task knowledge. Entanglement occurs when knowledge about distinct tasks is mixed together in shared circuits or representations, making it more likely that one task will interfere with another during decoding \citep{higgins_beta-vae_2017, pastrana2022DisentanglingVariationalAutoencoders, pan2025UnderstandingMitigatingOverrefusal} and is reflected in unintended emergent behaviors during post-training \citep{betley_emergent_2026}. The latter is particularly important for AI safety, where one major challenge during post-training is to suppress harmful behaviors without reducing helpfulness \citep{touvron2023Llama2Open}, especially in cases where representations which support aligned and misaligned behavior may be entangled. We discuss further relevant literature in \cref{app:extended_related_work}.

We study this question in two controlled settings. Firstly, we compare curricula in an ICL setting that pretrains two sub-tasks either in an imbalanced fashion (where the training data distribution is initially dominated by one sub-task) or in a balanced manner (both tasks are sampled uniformly). We then measure the ability of refusal fine-tuning to selectively suppress only one of the two sub-tasks. Surprisingly, we find that imbalanced curricula introduce substantially less overrefusal \citep{pan2025UnderstandingMitigatingOverrefusal, wang2024SurgicalCheapFlexible}. Using ablations and activation patching, we demonstrate that this occurs because imbalanced pretraining acquires ``disentangled'', task-specific circuits, whereas balanced training routes tasks through a shared pathway. Secondly, we design a synthetic language learning task in which sentences must satisfy a set of rule-based constraints, providing a controlled setting for studying alignment, in which aligned and misaligned behavior is defined by rule satisfaction or violation. Using the same balanced and imbalanced training regimes, we find that curricula which prioritize valid sentences early result in both more robust rule adherence and more localized rule representations. Together, these results reveal how curricula shape circuit- and network-level representations, and which kind of pretrained representations support selective fine-tuning. They also motivate the use of curricula during pretraining to facilitate downstream safety and alignment.

\textbf{We make three contributions}:
\begin{enumerate}
  \item We demonstrate that imbalanced curricula promote the selectivity of downstream refusal fine-tuning in a multi-task setting, with implications for safety and alignment.
  \item We explain this effect via interpretability analyses and show that imbalanced curricula induce disentangled, task-specific circuitry.
  \item We extend our investigation of curricula to a synthetic language learning setting where imbalanced curricula produce more localized rule representations and robust rule adherence.
\end{enumerate}

\section{Curricula in a controlled in-context learning task}
\label{sec:general_icl_heading}

\begin{figure}[h]
\centering
\includegraphics[width=1\textwidth]{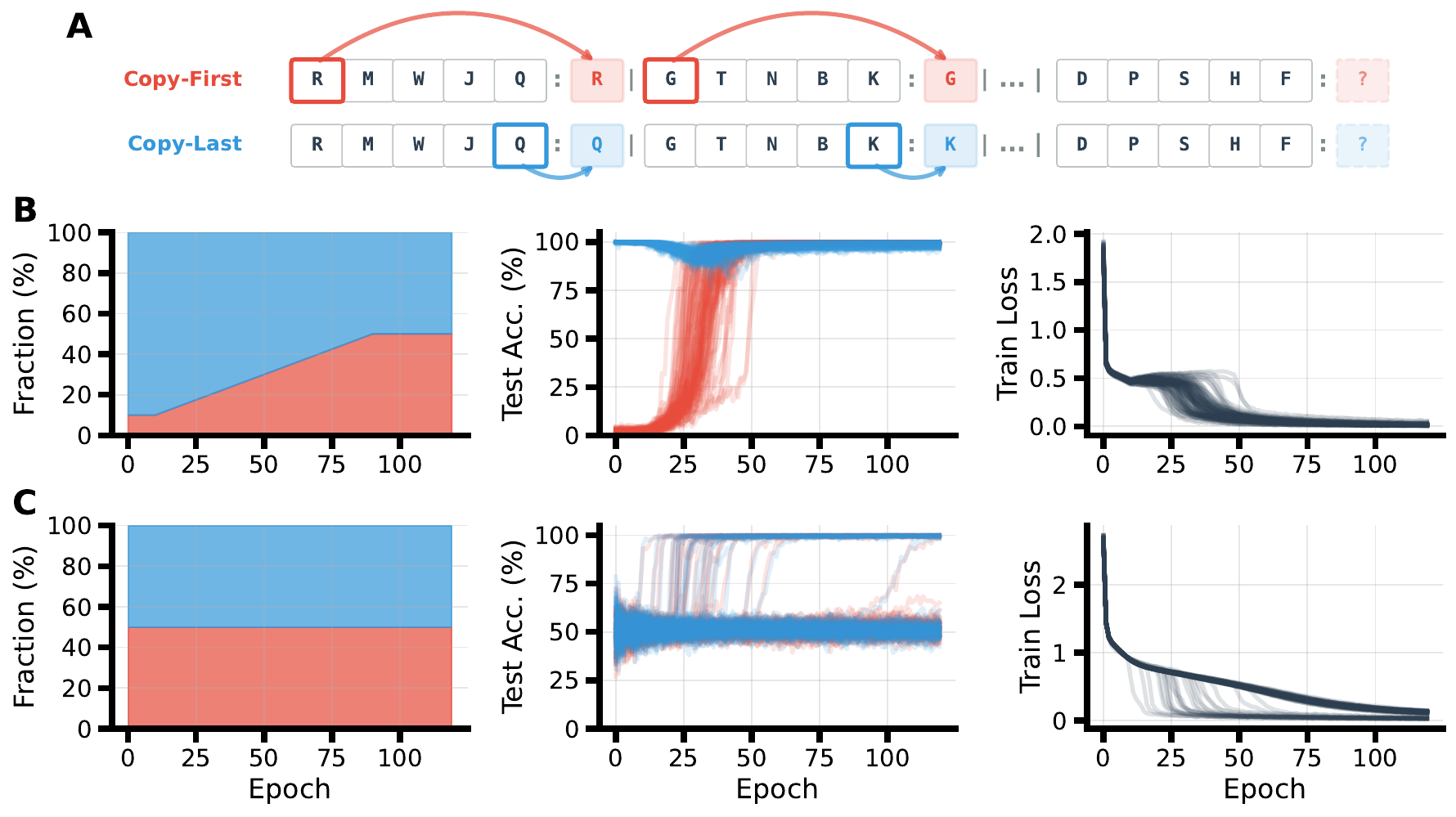}
\caption{\textbf{Task and training dynamics under imbalanced versus balanced.} (\textbf{A}) In the copy-first / copy-last task, each sequence contains $n=4$ labeled context examples followed by an unlabeled query. Within a sequence, all context examples share a latent rule: the label is either the first symbol (\emph{copy-first}) or the last symbol (\emph{copy-last}) of each example. To predict the query label, the model must infer the active rule from the context and apply it to the query. (\textbf{B}) Imbalanced training, in which the task mixture starts uneven and is gradually annealed to 50:50. Left: task fraction over training (red: copy-first, blue: copy-last). Center: held-out test accuracy on both tasks, lines show individual seeds. Right: training loss. Under the curriculum, all $100/100$ seeds converge to generalizing solutions, lines show individual seeds. \textbf{(C)} balanced training with a fixed 50:50 task mixture. While some seeds generalize, only $22/100$ reach high ($\geq90\%$) test accuracy. Remaining seeds achieve low training loss without corresponding test accuracy, consistent with a memorizing solution.
}
\label{fig:task_setup_and_training}
\end{figure}

\paragraph{Dataset and curricula}
\label{sec:icl_dataset}

We first study a task in which the network is required to copy either the first (\emph{copy-first} task) or last (\emph{copy-last} task) token from short strings consisting of $k = 5$ tokens. Sequences are a series of $n = 4$ ``context'' string-copy pairs, followed by a query string for which the to-be-copied item is not provided (see \cref{fig:task_setup_and_training}A). The network is not explicitly cued to perform copy-first or copy-last task but must infer whether to copy the first or last token based on the responses provided in the context sequence (\cref{fig:task_setup_and_training}A). This setup is an example of a ``context-dependent decision-making'' task \citep{flesch_orthogonal_2022, mante.etal2013a} and shares similarities with ICL ``task recognition" paradigms \citep{pan2023WhatInContextLearning, lin2024DualOperatingModes}.

Our key manipulation is the order in which copy-first and copy-last tasks are introduced during training. We focus on two curricula, which we label ``imbalanced'' and ``balanced''. In the \emph{imbalanced} curriculum, the relative probability of copy-last:copy-first is linearly interpolated from 90:10 to 50:50 over training (\cref{fig:task_setup_and_training}B, left). In the \emph{balanced} curriculum, examples of both tasks are presented with equal probability throughout training (\cref{fig:task_setup_and_training}C, left). Introducing the copy-first task early yields equivalent results (see \cref{app:copy_first_early}). For the main experiments, we use a \emph{fixed} training set of $S = 20{,}000$ sequences that the model revisits every epoch. Evaluation uses a held-out set of $500$ novel sequences sampled at a 50:50 task ratio (further details on curricula and data in \cref{app:icl_setup} and for experiments with an infinite number of sequences see \cref{app:infinite data regime}).

\paragraph{Training details}
\label{sec:icl_training_details}
We train a transformer network with two layers (L1 and L2) \citep{vaswani2017AttentionAllYou}\footnote{Upon acceptance of this paper, our code will be released at \url{https://github.com/kjsandbrink/alignment-dynamics.git}}.
Each layer contains a causal multi-head self-attention (ATT) block followed by a feed-forward network (FFN) of hidden dimension $4 \times d_{\text{model}} = 512$. Our base architecture uses $d_{\text{model}} = 128$ with $4$ attention heads per layer (head dimension $d_{\text{model}}/4 =32$), learned positional embeddings, and layer normalization. As in previous work on mechanistic interpretability \citep{gibson2026DistinctMechanismsUnderlying, singh2025StrategyCoopetitionExplains}, we chose this architecture because it suffices to produce our key phenomena while remaining amenable to detailed analysis. We use the AdamW optimizer \citep{loshchilov2019DecoupledWeightDecay} with $\beta_1 = 0.9$, $\beta_2 = 0.999$, a learning rate of $3 \times 10^{-4}$, and a batch size of $64$ sequences (see \cref{app:icl_setup} for additional details). We replicate our results for a wide range of settings, and for networks with for layers, in \cref{sec:hyperparameters}.

\subsection{Only imbalanced curricula reliably yield generalizing solutions}

Training under the two curricula results in differences in generalization performance. For the imbalanced curriculum, all $100/100$ seeds achieve low training loss and high test accuracy on new in-context examples for both sub-tasks by the end of training (\cref{fig:task_setup_and_training}B, center). Under balanced training, models quickly reach 50\% test accuracy corresponding to randomly copying either the first or the last token (See App. \cref{fig:icl-dynamics-behavior} for details). Even if training loss keeps decreasing for the models (\cref{fig:task_setup_and_training}C, right), only $22/100$ seeds further improve their test accuracy (\cref{fig:task_setup_and_training}C, center). This suggests that, for our setup, imbalanced curricula reliably help the model to generalize, whereas balanced curricula memorize more frequently. For imbalanced curricula, both tasks are learned one after another (see \cref{fig:task_setup_and_training}B center and App. \cref{fig:icl-dynamics-behavior}). On the other hand, for the balanced regime, models acquire both tasks at once (\cref{fig:task_setup_and_training}C center, for a calculation see App. \cref{fig:icl_learning_time_diff}). This generalization advantage of imbalanced curricula also generally holds across a range of hyperparameters including networks with 4 layers (see \cref{fig:hparam-fraction}).

\subsection{Imbalanced curricula reduce interference during refusal fine-tuning}
\label{sec:ICL_refusal}

In an analogy to alignment training, we next consider performance on a fine-tuning paradigm in which the model is trained to refuse to execute one, but not the other, of the two tasks. We ask whether there are any differences between models trained in the balanced or imbalanced curricula. On sequences of the \emph{refused} task, the original target is replaced with a dedicated refusal token. During fine-tuning, we include the unrefused (kept) task as 10\% of the sequences to avoid catastrophic forgetting. We evaluate on a held-out standard 50:50 (copy-first / copy-last) test set and track the refusal rate on both sub-tasks. In this analysis, we only include the subset of balanced seeds that also achieve high held-out accuracy on both sub-tasks ($\geq90\%$, $100/100$ seeds in the imbalanced and $22/100$ in the balanced condition survive this threshold). Additional implementation details are in \cref{app:icl_refusal} and for a replication of the main effects with more seeds see \label{app:replication_more_seeds}. 

\begin{figure}[t]
\centering
\includegraphics[width=1\textwidth]{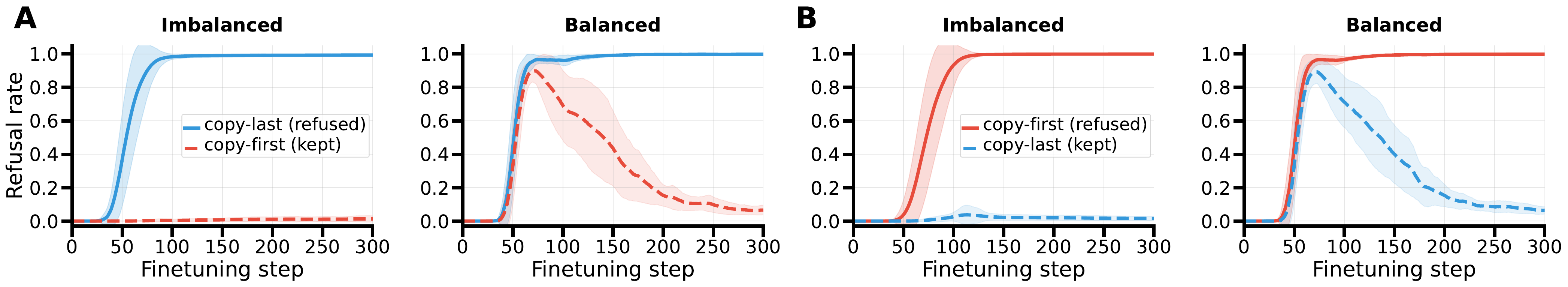}
\caption{\textbf{Refusal fine-tuning produces less overrefusal under imbalanced training.} Pretrained models are fine-tuned to emit a
refusal token in place of either the copy-last or copy-first target. We track the mean
refusal rate on held-out copy-last queries (solid, \emph{refused})
and on held-out copy-first queries (dashed, \emph{kept}); bands show standard deviation across seeds. (\textbf{A}) Left: For models trained with an imbalanced curriculum the copy-first refusal rate
remains low while copy-last is refused at a high rate. Right: Models trained in a balanced fashion show high interference, such that the model erroneously refuses the copy-first task during the early phase of refusal fine-tuning. (\textbf{B}) same as (A) but the model is trained to refuse the copy-first task our results hold for both types of refusal.}
\label{fig:refusal_rate}
\end{figure}

When learning to refuse either task, refusal fine-tuning successfully increases the refusal rate on held-out queries (see \ref{fig:refusal_rate}A for copy-last as the refused task, and \cref{fig:refusal_rate}B as copy-first as the refused task). However, the pretraining curriculum changes how refusal fine-tuning interferes with the other (kept) task. For models trained on imbalanced curricula, the overrefusal rate (i.e. rate of refusals on the unrefused task) remains low throughout fine-tuning (left subpanels). By contrast, balanced models exhibit substantial spillover (right subpanels). Further, the kept task accuracy does not degrade for imbalanced models (App. \cref{sfig:accuracy-refusal-training}). These results broadly hold across a range of hyperparameters (see \cref{fig:hparam-refusal} and \cref{fig:hparam-refusal-4layers}). Thus, refusal fine-tuning induces markedly more overrefusal in models trained under the balanced curriculum. Next, we study how differences in the pretrained model circuitry give rise to the observed differences in overrefusal.

\subsection{Imbalanced curricula display separated circuit organization}
\label{sec:circuit_analyses}

\paragraph{Ablations reveal differential component dependencies}
\label{sec:ablation_results_icl}

\begin{figure}[t]
\includegraphics[width=1\textwidth]{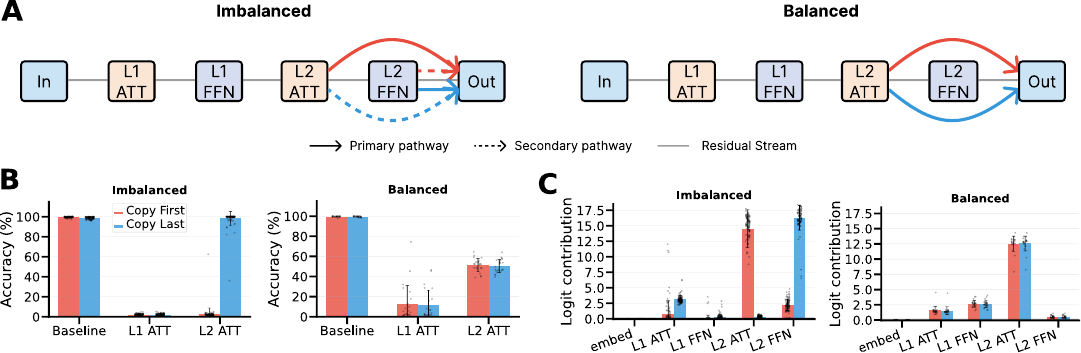}
\caption{\textbf{Imbalanced training produces task-specialized
sub-circuits, balanced training promotes a shared pathway.} (\textbf{A}) Illustration of how task-information flows through components of the transformer. Solid arrows mark each task's primary pathway, dashed arrows a secondary pathway. (\textit{left}) Under \emph{imbalanced} training, copy-first is routed primarily through L2~ATT with a secondary FFN pathway, while copy-last is routed primarily through L2~FFN with a secondary L2~ATT pathway. (\textit{right}) Under \emph{balanced} training, both tasks route through L2~attention alone, an entangled circuit. (\textbf{B})~Held-out accuracy after mean-ablating each
attention block (bars: mean\,$\pm$\,s.d.; dots: individual seeds).
Ablating L2~ATT eliminates copy-first under both conditions but
leaves copy-last intact in the imbalanced regime. (\textbf{C})~Direct
logit attribution of each component to the correct-token logit.
L2~FFN contributes strongly to copy-last for imbalanced training but is near
zero under balanced, while L2~ATT strongly contributes to the copy-first task.}
\label{fig:mechinterp_combined}
\end{figure}

To identify components needed for both learned sub-tasks, we start with mean-ablations \citep{wang2022InterpretabilityWildCircuit} of whole attention blocks after pretraining (see \cref{app:ablation_icl} for details). For models trained on balanced curricula (focusing on the seeds that generalize), mean-ablating the L2 attention block disrupts both copy-first and copy-last such that mean performance drops to $51.9\%$ and $42.9\%$ respectively (\cref{fig:mechinterp_combined}B, right). This is consistent with both sub-tasks relying on a shared attention-dependent circuit. For imbalanced curricula, however, ablating L2 attention only disrupts copy-first such that accuracy drops to $2.06\%$ while performance on copy-last is preserved at $99.1\%$ accuracy (\cref{fig:mechinterp_combined}B, left). Further, task identity is only strongly linearly represented in the residual stream of imbalanced models (App. \cref{sfig:linear_probing_analysis}). The second attention block is therefore not necessary to solve the copy-last task in this case, giving a first indication that imbalanced training induces more disentangled and task-specific circuitry.

\paragraph{Imbalance induces task-specific component contributions}
\label{sec:DLA_icl_results}

To quantify how strongly each model component contributes to the correct answer, we next use direct logit attribution. We project each model component's residual contribution onto the response's unembedding direction (DLA, \citep{elhage2021mathematical, wang2022InterpretabilityWildCircuit}, see \cref{app:dla_icl} for details). Larger positive attribution indicates that a component directly increases the logit of the correct answer. %
For models trained in the balanced setting, L2 attention contributes strongly to the correct answer logits for both the copy-first and copy-last task (\cref{fig:mechinterp_combined}C, right). In contrast, models trained on imbalanced curricula have highly task-specialized components, with L2 attention mostly contributing to the copy-first task and L2 FFN to the copy-last task \cref{fig:mechinterp_combined}(C, left). Thus, for our setting, DLA confirms that imbalanced training induces task-specialized downstream contributions, while interleaving routes both tasks through the same dominant component.

\paragraph{Causally localizing distinct circuits with activation patching}
\label{sec:patching}

While ablations and DLA identify which components are necessary and which contribute to the output, they do not causally reveal how task-relevant information flows through the network. We therefore use denoising activation patching to causally localize the pathways through which each sub-task is implemented \citep{heimersheim2024HowUseInterpret, meng2023LocatingEditingFactual}. We construct matched pairs of inputs, a \emph{clean} sequence sampled from one task and a \emph{corrupt} sequence in which the labels of the in-context examples are swapped to the other task's rule (the query label is held fixed). Clean sequences are solved at ceiling ($\geq 98\%$ accuracy); corrupt sequences drive the model to follow the rule that is indicated by the flipped in-context examples ($\leq 4\%$ accuracy on the original task). We then run a forward pass on the corrupt input but \emph{restore} a specific channel's activation from the clean run. We measure whether this suffices to recover the original task prediction using
$(\mathrm{Acc}_{\mathrm{patched}} -
\mathrm{Acc}_{\mathrm{corrupt}}) /
(\mathrm{Acc}_{\mathrm{clean}} - \mathrm{Acc}_{\mathrm{corrupt}})$, averaged across seeds. The channels we patch are the individual $\text{Q, K, V}$ matrices of L2's attention block and the L2 FFN residual read (see \cref{app:patching_icl} for complete protocol and full recovery table).

Under balanced training, both tasks are recovered entirely by restoring the L2 attention block's K-input and not at all by the L2 FFN (see \cref{fig:ICL-path_patching}A, right). Under imbalanced training, copy-first is recovered primarily by restoring L2 attention block's K-input and only partially by the L2 FFN. In contrast, copy-last shows an inverse pattern, recovering primarily by the L2 FFN and only partially by K (see \cref{fig:ICL-path_patching}A, left). This result converges with the ablation and DLA results, suggesting that in the imbalanced curriculum, the two tasks use different, albeit partially overlapping, circuits, while both tasks are solved via a single pathway under balanced training. \Cref{fig:mechinterp_combined}A summarizes our interpretability results. %
Together, these results suggest that models learn fundamentally more disentangled and task-specific circuitry in imbalanced curricula. We therefore hypothesize that entanglement of task circuitry drives overrefusal. Indeed, overrefusals appear to be significantly correlated with circuit specialization across a range of hyperparameters (\cref{fig:hparam-correlation}), including in rare settings in which balanced curricula lead to specialized task solutions. Strengthening this link, in the next section, we examine how refusal fine-tuning directly builds on existing circuitry. 

\subsection{Refusal fine-tuning co-opts existing model circuitry}
\label{sec:refusal_patching}

\begin{figure} [t]
\includegraphics[width=1\textwidth]{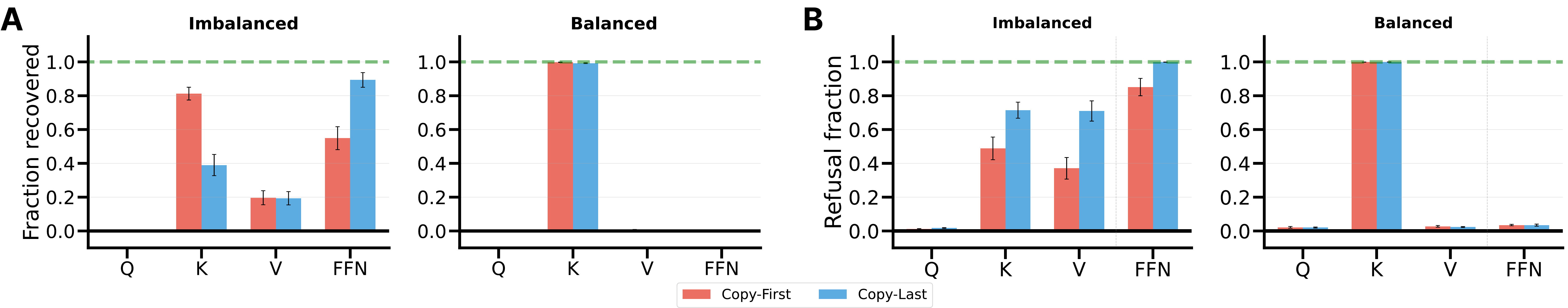}
\caption{\textbf{Activation patching reveals curriculum-dependent circuitry for pretrained and fine-tuned models.} (\textbf{A}) Fraction of held-out accuracy recovered when a single component is restored from a clean run into a corrupt run on the pretrained model. Restored channels: L2 attention projection inputs (Q/K/V), and L2 FFN, (bars show mean $\pm$ s.e. over seeds).
Under Imbalanced, copy-first localizes more strongly to L2 K and copy-last to the L2 FFN; under Balanced, L2 K alone recovers both tasks. (\textbf{B}) Activation patching on refusal-finetuned models. For each model, we take evaluation sequences for the task it was trained to refuse and construct matched counterparts by flipping the context labels to the opposite task. We cache activations from a forward pass on the original sequences, then patch individual L2 components into a forward pass with flipped labels and measure whether this induces refusal. Refusal is measured on (\textit{red}) copy-first sequences for a model refusal-trained on copy-last and on (\textit{blue}) copy-last sequences for a model refusal-trained on copy-first.
}
\label{fig:ICL-path_patching}
\end{figure}

To understand if differences during refusal training are directly connected to the above differences in model organization, we lastly investigate the circuitry in the refusal fine-tuned model. 
The L2 components contribute most strongly to the refusal logit for refused sequences (see \cref{fig:dla-refusal-balanced} for DLA analysis on the fine-tuned model). 
We therefore use activation patching on the fine-tuned model's L2 components to investigate the circuitry causally. We again use matched pairs of inputs from copy-first and copy-last tasks, which have flipped context labels but are otherwise identical. For models trained to refuse copy-first, we selectively patch individual model components cached from a copy-first into a copy-last forward pass. We then measure which component patches lead models to flip their answer to refuse. For models trained under the balanced curriculum, for both tasks, patching the L2 attention block's K-input drives refusal behavior while the L2 FFN does not contribute (\cref{fig:ICL-path_patching}B, right). In contrast, under the imbalanced curriculum we find that both channels of the L2 attention block (K\&V) as well as L2 FFN contribute to refusal (\cref{fig:ICL-path_patching}B, left). Therefore, for both conditions, refusal appears to be implemented via those components that were most important in the original circuit (after pretraining) rather than, for instance, selectively modifying the readout. We further show that task directions in the model residual stream \cref{sfig:cos_dist_of_refuse_v_kept} are more orthogonal in imbalanced models. Thus, for balanced curricula overrefusal appears to be caused by shared mechanisms which are co-opted during fine-tuning. %

\section{Curricula in a synthetic language learning task}

To study whether the curriculum effects in the previous experiment scale to a richer setting, we train the model on a synthetic language learning (SLL) task characterized by a grammar and a set of prioritized conditional rules. We define ``aligned'' sentences as those that satisfy all applicable rules (note that ``aligned'' and ``misaligned'' refer specifically to rule satisfaction within the SLL task and do not make broader claims about AI alignment). This provides a controlled framework for specifying rule-based norms. Similar to the previous section, we train models under balanced and imbalanced mixtures of aligned and misaligned data. Mirroring real-world alignment pipelines, we then fine-tune them on only aligned data. This setup allows us to probe whether curricula encourage the formation of representations that disentangle rule-consistent from rule-violating behavior, and whether this influences the effectiveness of downstream fine-tuning.

\begin{figure}[ht]
\centering
\includegraphics[width=1\textwidth]{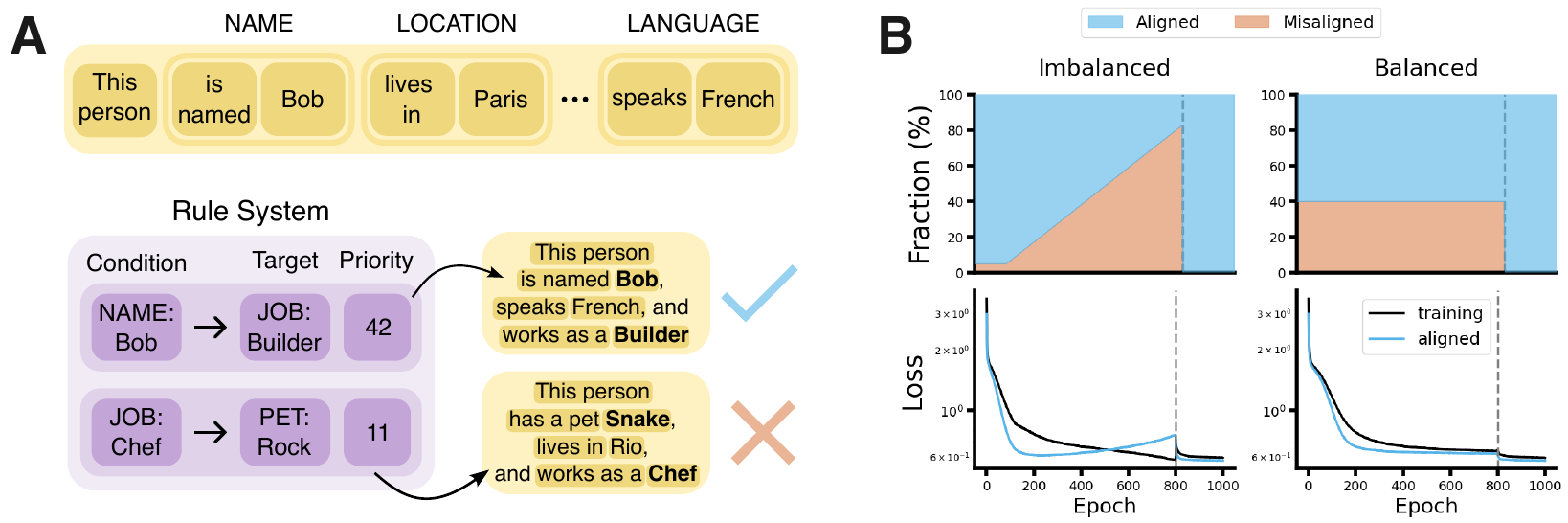}
\caption{\textbf{Synthetic language learning (SLL) task for studying rule-based alignment under balanced and imbalanced curricula}. 
        (\textbf{A}) Top: The SLL task operates on biographical profiles with features (e.g. JOB, LOCATION, LANGUAGE) taking specific values (e.g. Bob, Paris, French).
        Bottom: The rule system specifies condition values which, when met, specify the value for the target feature (Name: Bob $\rightarrow$ Job: Builder). 
        (\textbf{B}) Top: We compare two curricula: (1) Balanced, maintaining a constant level of misaligned data. (2) Imbalanced, beginning with low levels of misalignment and linearly increasing it over training (same overall exposure). Pretraining is followed by a fully aligned fine-tuning (after the gray dotted line). Bottom: Mean loss on the full training data (dark blue) and the aligned subset (light blue), shaded regions indicating $95$\% CI are included but not visible.}
\label{fig:p2_task_setup}
\end{figure}

\paragraph{Synthetic language dataset}
The dataset consists of a combinatorially large set of sentences describing short profiles, with features such as NAME, JOB, and LOCATION (see \cref{fig:p2_task_setup}A top). This is inspired by previous work on learning dynamics in transformers \citep{allen-zhu2024PhysicsLanguageModels, zucchet.etal2025}. These sentences follow a grammatical template, providing syntactic structure for the model to learn (see \cref{app:sll_setup}). On top of this grammar, we define a set of rules which constrain which combinations of features are permissible (or ``aligned''), see \cref{fig:p2_task_setup}A bottom. Each rule specifies a condition (e.g. NAME = Bob) and a target (e.g. JOB = Builder). When the condition is met, the corresponding target must also hold. Higher-priority rules take precedence when multiple rules apply. We employ a similar training setup to the previous section, training 20 seeds per curriculum, with full details in \cref{app:dlr_architecture}. 

\subsection{Learning dynamics across curricula}

As in the previous task we compare two training regimes: (1) \emph{imbalanced}, where the proportion of misaligned samples begins low and linearly increases across training (from 5\% to 80\%), and (2) \emph{balanced}, where the misaligned proportion remains fixed (at 40\%) throughout (additional curricula are discussed in \cref{app:sll_extra_curricula}). These conditions, importantly, contain the same amount of misaligned data, but differ in how they are distributed across training time. Pretraining is followed by a fine-tuning phase on purely aligned samples, mimicking standard alignment fine-tuning pipelines. Model performance improves steadily in both conditions (\cref{fig:p2_task_setup}B). Under the imbalanced curriculum, loss on the aligned subset of training data initially falls and then increases as the proportion of misaligned data rises. Despite different pretraining curricula, the fine-tuning loss trajectories are similar.

\subsection{Surface-level vs. latent alignment behavior}

\begin{figure}[ht]
\centering
\includegraphics[width=1\textwidth]{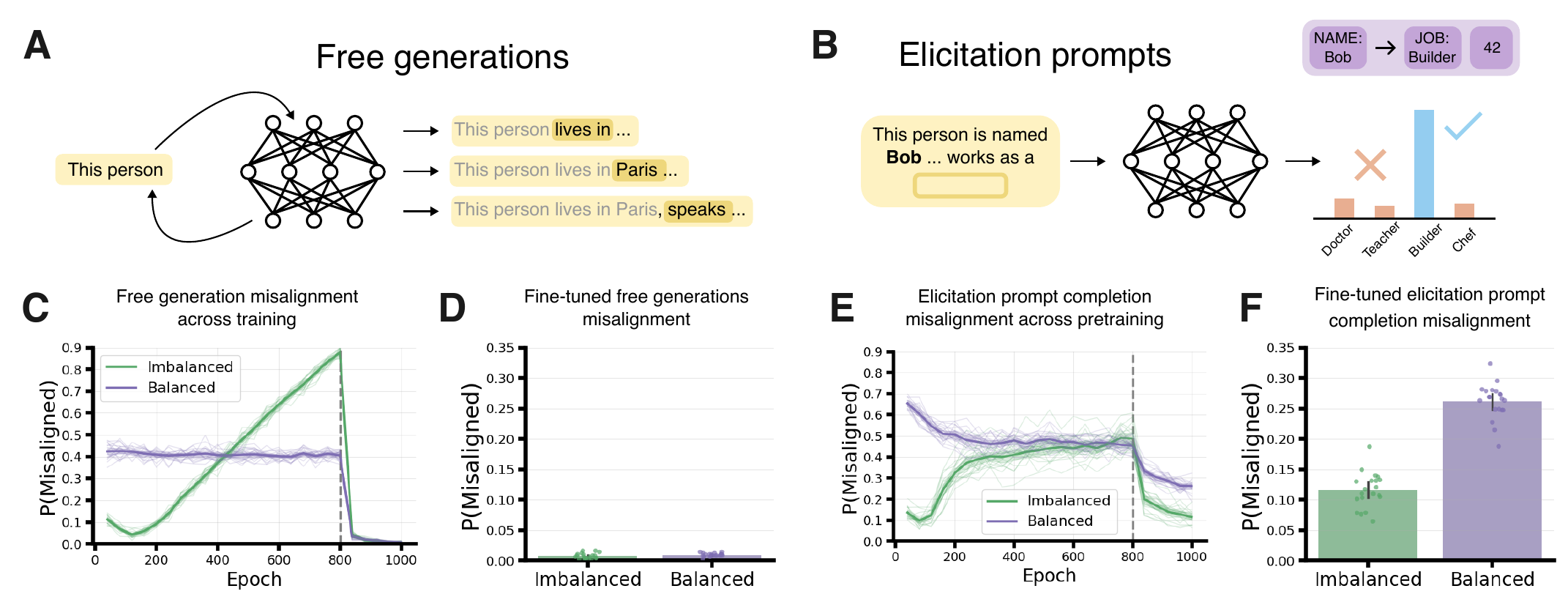}
\caption{\textbf{Targeted prompts reveal robustness of rule adherence.} 
        (\textbf{A}) Free generations: sequences auto-regressively sampled from the model.
        (\textbf{B}) Elicitation prompts: rule-triggering prompts where the final token queries the target feature, allowing a direct assessment of $\text{P}(\text{Misaligned})$.
        (\textbf{C}) $\text{P}(\text{Misaligned})$ of free generations across epochs under the imbalanced (green) and balanced (purple) regimes (thin: individual seeds; bold: mean; dashed line: start of fine-tuning).
        (\textbf{D}) After fine-tuning, mean $\text{P}(\text{Misaligned})$ is close to zero for both curricula (error bars: $95$\% CI over seeds; individual seeds: dots).
        (\textbf{E, F}) $\text{P}(\text{Misaligned})$ for elicitation prompt completions across training and after fine-tuning, where imbalanced curricula yield lower $\text{P}(\text{Misaligned})$ than balanced (same plotting conventions as C, D).
        }
\label{fig:p2_metrics}
\end{figure}

We assess ``model alignment'' using two complementary measures. \textbf{(i) Free generations}: 500 sentences sampled autoregressively from the model (\cref{fig:p2_metrics}A). \textbf{(ii) Elicitation prompts}: 250 sentences designed to trigger a specific rule by satisfying its condition. The final token is constructed to query the target feature, allowing for a direct readout of the probability of misaligned completions (\cref{fig:p2_metrics}B). Free generations probe sampled autoregressive behavior, while elicitation prompts directly probe the model's conditional next-token distribution in controlled rule-triggering contexts. For both measures we evaluate whether the output sentences are grammatically valid (\cref{app:sll_beh_grammar}) and, if so, measure the probability of producing a misaligned completion, $\text{P}(\text{Misaligned})$.

\textbf{Free generations } \:Under both conditions, free generations largely reflect the statistics of the training data. Under balanced training, $\text{P}(\text{Misaligned})$ remains approximately constant at $\sim0.4$, while under imbalanced $\text{P}(\text{Misaligned})$ steadily increases as the proportion of misaligned training data grows (\cref{fig:p2_metrics}C). During fine-tuning, $\text{P}(\text{Misaligned})$ drops close to zero for both conditions, with negligible differences between them at the end of fine-tuning (\cref{fig:p2_metrics}D). This suggests that fine-tuning is similarly effective across training regimes at the level of surface behavior.

\textbf{Elicitation prompts } \:In contrast, elicitation prompts reveal clear differences between curricula. At the end of pretraining, imbalanced and balanced reach similar levels of $\text{P}(\text{Misaligned})$, see \cref{fig:p2_metrics}E. However, after fine-tuning, the models trained under the imbalanced curriculum perform substantially better at concentrating probability mass on aligned completions, exhibiting significantly lower $\text{P}(\text{Misaligned})$, see \cref{fig:p2_metrics}F. This indicates that imbalanced training yields representations that are more selectively aligned during fine-tuning. We further establish the robustness of this effect on a range of rule systems in \cref{app:sll_rulesys_robust}.

Taken together, these results reveal differences between surface-level behavior and underlying rule representations. While both curricula produce well-aligned free generations after fine-tuning, elicitation prompts reveal that models trained under the balanced curriculum exhibit a higher probability of producing misaligned completions when directly probed (additional curricula in App. \cref{fig:sll-control-curricula}). This indicates that seemingly well-behaved models may still fail to robustly internalize the underlying rule structure, and that curricula improve robustness of alignment under targeted evaluation.

\subsection{Early-layer attention drives rule adherence}

\paragraph{Early-layer attention to condition-tokens}

\begin{figure}[ht]
\centering
\includegraphics[width=1\textwidth]{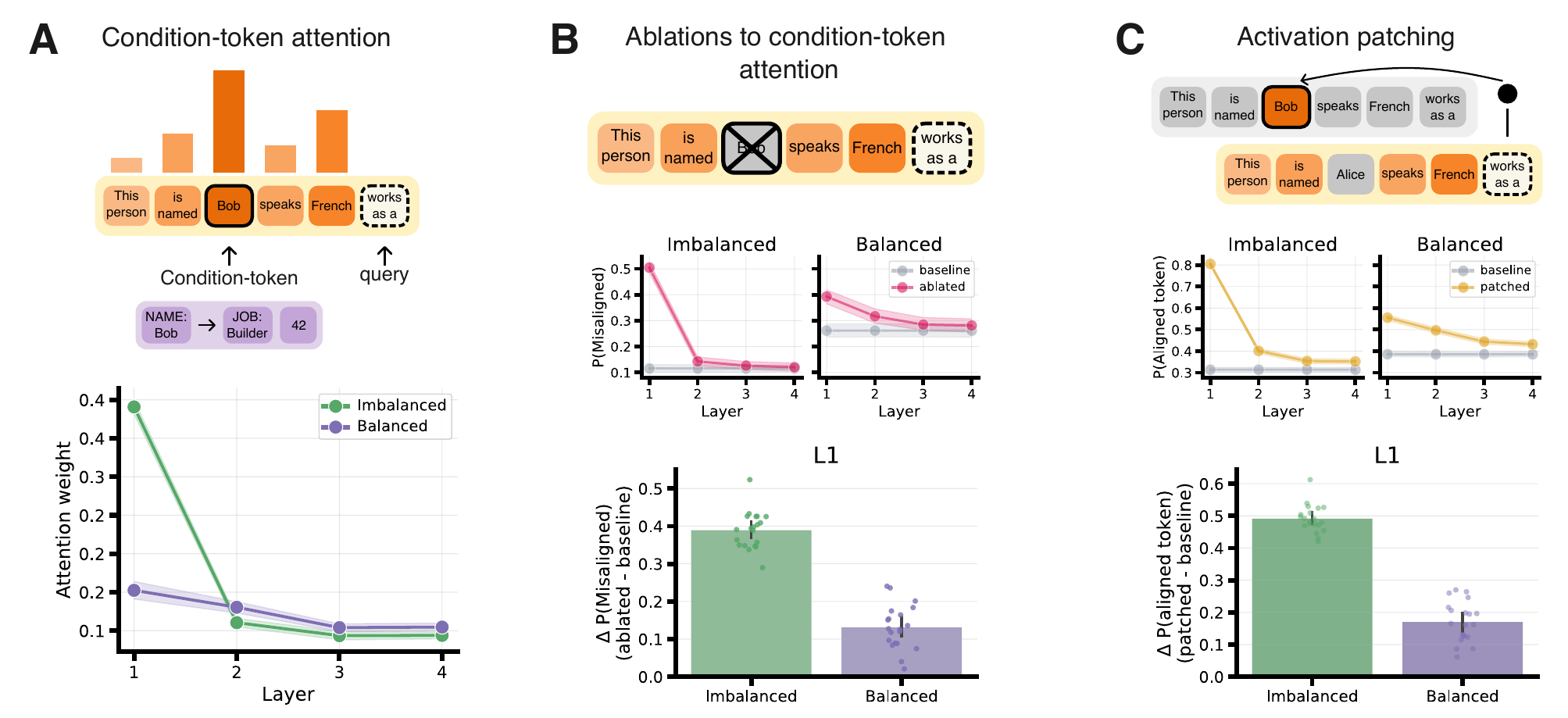}
    \caption{\textbf{First layer attention is critical for rule adherence, especially under imbalanced curriculum.} 
    (\textbf{A})
    Top: Schematic of attention from final token to preceding tokens. Condition-token (``Bob'') specifies that a given rule applies.
    Bottom: Mean attention weight to condition-token across layers of transformer for imbalanced (green) and balanced (purple) models (shaded region indicates $95$\% CI, for all panels).
    (\textbf{B}) Top: Ablation of condition-token.
    Bottom: $\text{P}(\text{Misaligned})$ under ablation relative to baseline across layers (balanced vs. imbalanced). A high probability for misaligned output after ablation reveals that this attention weight is critical to rule-adherence.
    (\textbf{C})
    Top: Activation patching from a rule-triggering source (top) sequence to a rule-neutral destination (bottom) sequence.
    Bottom: Probability of producing the completion that would have been required by the source sequence ($\text{P}(\text{Source target})$) for patched models vs. baseline across layers. Patched values above baseline indicate that this specific attention block output drives rule-adherence.
    }
\label{fig:rule_mech_interp}
\end{figure}

To understand how imbalance leads to more robust rule-following, we analyze how fine-tuned models integrate condition information during inference. In the elicitation prompts, the final token (e.g. ``works as a''), queries the target feature value, and the model must incorporate information from earlier tokens that determine whether a rule applies (e.g. the person's name is ``Bob'', we refer to these as \emph{condition-tokens}). In transformers, communication between tokens occurs via attention. Therefore, for each layer, we extract the attention-weight from the query-token to the condition-token. 
\Cref{fig:rule_mech_interp}A shows that models across both curricula exhibit greater attention to condition-tokens in early layers. Crucially, at L1, condition-token attention is significantly greater under imbalanced training relative to balanced.%

\paragraph{Ablation of condition-token attention}

We next performed targeted attention ablations to determine the causal contribution of condition-token attention at each layer. For a given layer, we zero the attention 
from the final query-token to the condition-token position and renormalize the attention distribution (details in \cref{app:sll_ablation_details}). This intervention targets a single activation while minimally perturbing the rest of the attention pattern. Across both training regimes, ablating early-layer condition-token attention leads to an increase in $\text{P}(\text{Misaligned})$ relative to baseline (\cref{fig:rule_mech_interp}B), with negligible effect on grammatical correctness (\cref{app:sll-grammar}). Later-layer ablations have progressively smaller effects, suggesting condition-token information is primarily integrated at early layers. Crucially, at L1, the increase in $\text{P}(\text{Misaligned})$ is substantially greater under imbalanced relative to balanced, confirming a stronger causal dependence on early-layer attention for rule-following.

\paragraph{Activation patching of final-token representations}

To test whether condition-token information in early layers is not just necessary, but sufficient for prediction, we perform activation patching at the final token \citep{heimersheim2024HowUseInterpret}. We replace the attention block output at the final token for a given layer for a rule-neutral destination sentence with the corresponding activation from a rule-triggering source prompt, \cref{fig:rule_mech_interp}C, and measure whether this intervention steers the model's prediction toward the target completion from the source prompt ($\text{P}(\text{Source target})$, details in \cref{app:sll_patching_details}). 
Across both balanced and imbalanced training, patching early-layer activations substantially increases $\text{P}(\text{Source target})$ relative to baseline, with smaller effects at later layers (\cref{fig:rule_mech_interp}C). While this intervention does not isolate condition information alone, it demonstrates that the representations formed at the first layer are sufficient to drive rule-consistent predictions, suggesting that condition-token information is processed in this layer. Importantly, the effect of patching L1 attention outputs is significantly greater for imbalanced compared to balanced, indicating that models trained under this regime rely more strongly on L1 representations to encode condition-dependent behavior.

Taken together, these results show that curricula influence both the robustness of rule-following behavior and the internal organization of rule representations. While both curricula yield similarly well-aligned surface-level behavior after fine-tuning, elicitation prompts reveal that models trained under balanced curricula are more prone to produce misaligned completions when directly probed. Mechanistically, this difference is reflected in how condition information is processed. Imbalanced training leads to more localized processing of condition information, with a strong dependence on first-layer attention (which converges faster than other layers, \cref{fig:sll-qkv-weight-dynamics}), in contrast to balanced training which distributes this computation across layers. These findings indicate that curricula shape not only model behavior, but the layer-wise organization of rule-relevant information.%

\section{Discussion and Conclusion}

We have shown that curricula induce differences in learning dynamics, generalization, and model circuitry. These differences have important downstream consequences: For balanced curricula models are more likely to overrefuse during selective fine-tuning. Similarly, in the SLL task that includes notions of ``aligned'' and ``misaligned'' sentences, the imbalanced curriculum enabled the formation of more localized attention patterns, resulting in more effective fine-tuning for rule adherence. Prior work has identified that burstiness within sequences and Zipfian distributions over examples contribute to the emergence of ICL \citep{chan_data_2022, reddy_mechanistic_2023, singh2023TransientNatureEmergent}. We similarly find that imbalanced distributions 
over tasks can drive ICL abilities and that learned model circuitry can differ as a function of such curricula. This identifies curricula as a potential factor in driving the formation of shared circuitry as described in the mechanistic interpretability literature \citep{merullo_circuit_2024}. LLMs mostly display overrefusal when models struggle to distinguish between subtle differences between benign and harmful content \citep{pan2023WhatInContextLearning, wang2024SurgicalCheapFlexible, touvron2023Llama2Open}. Our results indicate that curriculum-driven circuit entanglement may modulate this effect. %
verrefusal for tasks that were trained simultaneously, and hints at the role of curricula and learning dynamics in these phenomena. 

While our results show curriculum phenomena in two settings, the extent to which these phenomena are present in larger models, which have to learn a wide array of (implicit) sub-tasks, is not yet clear. There is evidence in LLMs that the implicit curricula relate to task representations \citep{liu2026WhatLanguageModels}. Similarly, the simplified nature of our task settings potentially limits their relevance to phenomena in ethological settings. Future work should explore how selective fine-tuning differentially affects tasks with similar representations in LLMs, and how explicit pretraining curricula change learned circuitry. 
Our work has important consequences for pretraining: Practitioners should take the dynamics of learning into account and be aware that interference during fine-tuning can be greater for tasks trained in tandem.

\section{Acknowledgments}
We would like to thank Edmund Lau, our research sponsor, for his valuable guidance, feedback and support. We also thank Andrew Saxe, Naomi Saphra, Robert West, Julian Minder and Viktor Moskvoretskii, for helpful and insightful discussions. SAB, JR, KJS, and MHW were supported by the UK AI Security Institute (AISI) through an Alignment Project grant (AP-S2-100060).
CS was supported by a Wellcome Trust Discovery Award (227928/Z/23/Z).

\newpage
\bibliographystyle{abbrvnat}
\bibliography{jirko_refs,align_refs,manual_refs}

\newpage
\appendix

\clearpage
\section*{Appendix Roadmap}
\addcontentsline{toc}{section}{Appendix Roadmap}

The appendix is organized as follows:

\begin{longtable}{@{}p{0.30\textwidth}p{0.62\textwidth}@{}}
\toprule
\textbf{Section} & \textbf{Contents} \\
\midrule
\endhead
\Cref{app:extended_related_work} & Extended related work: data-distributional properties in ICL, model circuits/mech.\ interp., refusal fine-tuning, curriculum learning, disentangled representations. \\[4pt]
\Cref{app:icl_setup} & ICL task setup: dataset/task details, architecture \& optimization, compute, pretraining curricula, refusal fine-tuning protocol. \\[4pt]
\Cref{app:icl_additional figures} & Additional ICL results: learning-time differences, policy dynamics, linear probing of task identity, entanglement over fine-tuning, copy-first-early curriculum, refusal accuracy, constant-imbalance curricula, infinite-data regime. \\[4pt]
\Cref{app:mechanism_icl} & Interpretability methods for the ICL task: mean-component ablation, direct logit attribution, activation patching (incl.\ full recovery table). \\[4pt]
\Cref{app:refusal_mechanism} & Mechanistic analyses of the refusal fine-tuned model (DLA on original vs.\ refusal token). \\[4pt]
\Cref{app:sll_setup} & SLL task setup: grammar, rule system, dataset construction, curricula, architecture \& optimization, compute. \\[4pt]
\Cref{app:sll_extra} & Additional SLL results: control curricula, grammaticality checks, robustness across rule-system complexity/dataset size. \\[4pt]
\Cref{app:ex_ana_ablation_patching} & Mechanism analyses for SLL: attention ablations, activation patching, dynamics across training, grammaticality under intervention, attention-weight dynamics. \\[4pt]
\Cref{sec:hyperparameters} & Hyperparameter sweep for the ICL task (learning rate, batch size, weight decay, width, imbalance fraction; 2- and 4-layer networks). \\[4pt]
\Cref{app:sll_hparam} & Hyperparameter sweep for the SLL task. \\
\bottomrule
\end{longtable}

\section{Extended Related Work}
\label{app:extended_related_work}

\paragraph{Data distributional properties in ICL}
\label{app:data_distributional_props}
Previous work has shown that properties of the data distribution, such as burstiness, the frequency of occurrence of rare classes, and Zipfian data distributions are crucial for the emergence of in-context learning abilities \citep{chan_data_2022, reddy_mechanistic_2023, pesnotlerousseau2026SharedSensitivityData}. Controlled function-learning studies further show that transformers can acquire in-context learning algorithms when
trained on suitable distributions over tasks or functions
\citep{garg2023WhatCanTransformers}. However, how the emergence and structure of model circuits differ as a function of learning curricula has not been explored in much detail.

\paragraph{Model circuits in mechanistic interpretability}
\label{app:mechinterp}
Mechanistic interpretability seeks to identify the circuits responsible for specific model behaviors \citep{hanna2023HowDoesGPT2, conmy2023AutomatedCircuitDiscovery, lieberum2023DoesCircuitAnalysis, wang2022InterpretabilityWildCircuit}. For example, the mechanisms underlying induction circuits \citep{olsson2022IncontextLearningInduction} in simple attention-based models are now understood in considerable detail \citep{reddy_mechanistic_2023, singh2024WhatNeedsGo, singh2025StrategyCoopetitionExplains, elhage2021mathematical}. 
At the same time, circuits initially thought to be task-specific can be gainfully reused across distinct tasks \citep{merullo_circuit_2024}. However, when and why models reuse circuitry across sub-tasks, and when they instead develop separated, sub-task specific circuitry \citep{michaud2023QuantizationModelNeural, quirke2024UnderstandingAdditionTransformers}, remains poorly understood. We propose that curricula, implicit or explicit, are one driver of whether sub-tasks come to share mechanisms or are implemented by more separated computations.

\paragraph{Refusal fine-tuning in LLMs}
\label{app:refusal_finetuning}
The alignment of LLMs for helpfulness and harmlessness is typically implemented via a pipeline that includes supervised fine-tuning and preference optimization such as reinforcement learning from human feedback and direct preference optimization \citep{ouyang_training_2022, bai2022TrainingHelpfulHarmless, rafailov2024DirectPreferenceOptimization}. In Supervised safety fine-tuning, models are directly trained on adversarial prompts with corresponding safe demonstrations \citep{touvron2023Llama2Open,xueRefusalTriggers2026, qi_safety_2024}. However, overrefusal where LLMs erroneously refuse benign queries can in some cases undermine the practical utility of these techniques \citep{touvron2023Llama2Open, xueRefusalTriggers2026}. The reasons for such overrefusal have been explored from several angles. For example, previous work has suggested these behaviors are likely when prompts are close to safety decision boundaries where models struggle to distinguish between harmful and harmless user requests \citep{pan2025UnderstandingMitigatingOverrefusal} and have been found to be induced by specific linguistic cues \citep{xueRefusalTriggers2026}. Mitigation techniques include the ablation of "overrefusal directions" \citep{wang2024SurgicalCheapFlexible} or through the targeted prompting with specific safety reflections \citep{si_think_2025}. While previous work has focused on representational factors driving overrefusal, our work seeks to explain how learning dynamics and curricula can give rise to entangled task representations that can give rise to unintended overgeneralization behaviors.

\paragraph{Curriculum learning}

Human and animal learning is greatly aided by the ordered presentation of information \citep{krueger2009FlexibleShapingHow, 1959-07839-001} and there have been considerable efforts to translate this benefit into techniques for machine learning \citep{bengio_curriculum_2009}. In spite of this, curriculum learning has proven non-trivial to successfully implement for neural networks \citep{warstadt2023babylm, wu_when_2021} and has not become an established part of the training pipeline of frontier LLMs \citep{zhang2026RandomSamplingEfficient}. Classic curriculum learning proceeds by gradually increasing the difficulty of presented training examples, which means it requires: (i) a difficulty criterion for ranking examples by difficulty, and (ii) a scheduler, for determining when an example of which difficulty level should be shown \citep{Soviany_2021_curriculum}. While superficially straightforward, this hides deep questions such as how to define difficulty, whether it is objective or student dependent, how the scheduler should manage sample diversity, and how it should space training samples to minimize forgetting. Curricula are therefore a tantalizing tool, but there are no generally reliable guidelines for their implementation yet.

\paragraph{Disentangled representations}
\label{app:dient_repr}

Disentangled representations are those that decompose factors of variation, i.e. they disentangle the individual elements of the environment or task, rather than representing multiple components in a combined, entangled manner \citep{bengio2014representationlearningreviewnew}. Prior work has proposed that disentangled representations support learning success, generalization \citep{steenbrugge_improving_2018}, robustness, and interpretability \citep{Lake_Ullman_Tenenbaum_Gershman_2017, higgins_beta-vae_2017}, and reduce the impact of catastrophic forgetting \citep{kirkpatrick_overcoming_catastrohpic}. \citet{locatello_challenging} demonstrated that disentangled representations cannot be reliably learned in fully unsupervised settings without inductive priors or supervision signals, suggesting that training structure may play a role in encouraging disentanglement. Humans preferentially form disentangled representations when trained in a blocked manner, i.e. focusing on a single task for an extended period \citep{flesch_comparing_2018, flesch_orthogonal_2022}. Finding curricula which promote disentangled representations for neural networks remains an area of active research.

\clearpage
\section{Experimental setup for the ICL task}
\label{app:icl_setup}

\subsection{Task and dataset}
\label{app:icl_task}

Each example string is a combination $k = 5$ tokens (drawn i.i.d.\ with replacement from a $50$-symbol vocabulary), followed by a colon, a single-symbol label, and a separator. A training sequence concatenates $n = 4$ labeled context examples followed by an unlabeled query (feature + colon); cross-entropy is applied at the query colon. Total sequence length is $38$ tokens. The vocabulary has $53$ tokens ($50$ symbols + colon + separator + a reserved refusal token). Each sequence is assigned a single task by a per-sequence Bernoulli draw and all $n+1$ examples follow that task; we resample features within a sequence until all $n+1$ are distinct. The training set is fixed at $K = 20{,}000$ sequences revisited each
epoch. The evaluation set is $500$ sequences sampled at $50/50$ with a different seed.

\begin{table}[h]
\centering
\small
\setlength{\tabcolsep}{4pt}
\renewcommand{\arraystretch}{1.15}
\begin{tabular}{l l l c}
\toprule
Task & Context examples & Query & Target \\
\midrule
\multirow{3}{*}{Copy-first (CF)}
  & \texttt{kqzbn:\textbf{k}\,|\,fhdsi:\textbf{f}\,|\,owrxc:\textbf{o}\,|\,vutpl:\textbf{v}}
  & \texttt{jaegm:} & \texttt{\textbf{j}} \\
  & \texttt{tymzq:\textbf{t}\,|\,leohr:\textbf{l}\,|\,dwnub:\textbf{d}\,|\,pivfg:\textbf{p}}
  & \texttt{ckxas:} & \texttt{\textbf{c}} \\
  & \texttt{rxhqj:\textbf{r}\,|\,mceyk:\textbf{m}\,|\,bdozi:\textbf{b}\,|\,sfglz:\textbf{s}}
  & \texttt{nvtwo:} & \texttt{\textbf{n}} \\
\midrule
\multirow{3}{*}{Copy-last (CL)}
  & \texttt{hzbyk:\textbf{k}\,|\,rcweo:\textbf{o}\,|\,pqxnu:\textbf{u}\,|\,vgmds:\textbf{s}}
  & \texttt{fjlta:} & \texttt{\textbf{a}} \\
  & \texttt{oebmq:\textbf{q}\,|\,ucpiv:\textbf{v}\,|\,nrdsa:\textbf{a}\,|\,ywhpz:\textbf{z}}
  & \texttt{gxlkt:} & \texttt{\textbf{t}} \\
  & \texttt{tphwq:\textbf{q}\,|\,jvbom:\textbf{m}\,|\,keurd:\textbf{d}\,|\,scfan:\textbf{n}}
  & \texttt{yzigl:} & \texttt{\textbf{l}} \\
\bottomrule
\end{tabular}
\vspace{7pt}
\caption{Illustration of example sequences for each task. Each sequence concatenates
$n{=}4$ in-context examples followed by an unlabeled query. An
example is a feature of $k{=}5$ symbols, a colon, a label symbol, and
a separator (\texttt{|}). Under \emph{copy-first} the label is the
first symbol of the query; under \emph{copy-last} the label is the
last symbol. The model is trained to predict the correct label at the
colon following the query (rightmost column). Symbols are drawn from
a $50$-symbol token vocabulary}
\label{tab:task_examples}
\end{table}

\subsection{Architecture and optimization}
\label{app:icl_architecture}

We use a $2$-layer Transformer with $d_{\text{model}} = 128$, with
$4$ heads per layer of dimension $d_{\text{model}}/4 =32$, GELU FFN of hidden dimension
$4\,d_{\text{model}} = 512$, learned positional embeddings and dropout $0.1$ \citep{srivastava2014DropoutSimpleWay}. Pretraining: AdamW \citep{loshchilov2019DecoupledWeightDecay},
$\beta_1{=}0.9$, $\beta_2{=}0.999$, lr $3\times10^{-4}$, weight decay
$0.01$, gradient-norm clipping at $1.0$, batch size $64$. All models were trained using PyTorch. 

\subsection{Compute}

We used 3 NVIDIA GeForce RTX 3090 GPUs for this study. For the main experiments in \cref{sec:general_icl_heading} training a single seed and a single curriculum for experiment 1 took around 20 minutes of GPU time, leading to a total compute amount of 30 hours (30 seeds for 3 curricula). %

\subsection{Pretraining curricula}
\label{app:icl_pretraining}

Three conditions: (i) imbalanced, copy-last early
(10:90); (ii) balanced (fixed at 50:50 task mixture). For imbalanced conditions, the copy-first fraction is held at the starting value for the first $10$ epochs of pretraining, linearly faded to 50:50 between $10$ and $90$ epochs, and held at $50\%$ thereafter for 30 epochs. We train $30$ seeds per condition. Analyses comparing curricula are restricted to seeds achieving $\geq 90\%$ held-out accuracy on both tasks.

\subsection{Refusal fine-tuning}
\label{app:icl_refusal}

Starting from the final pretrained checkpoint, we fine-tune on a $90\%$ refused-task / $10\%$ kept-task mixture: refused-task labels to the final query are replaced with the refusal token, kept-task targets to the final query are unchanged. AdamW with lr $5\times10^{-5}$, $3$ epochs over a fresh $20{,}000$-sequence training set (distinct seed from pretraining); other hyperparameters match pretraining. We evaluate the models on a 50:50 evaluation. We report refusal rate (fraction of last-position predictions equal to the refusal token) and kept task accuracy computed per task.

\section{Additional results for the controlled in-context learning task}
\label{app:icl_additional figures}

\subsection{Learning time difference between curricula}
\label{app:learning_time_diff}
\begin{figure} [h]
\centering
\includegraphics[width=.35\textwidth]{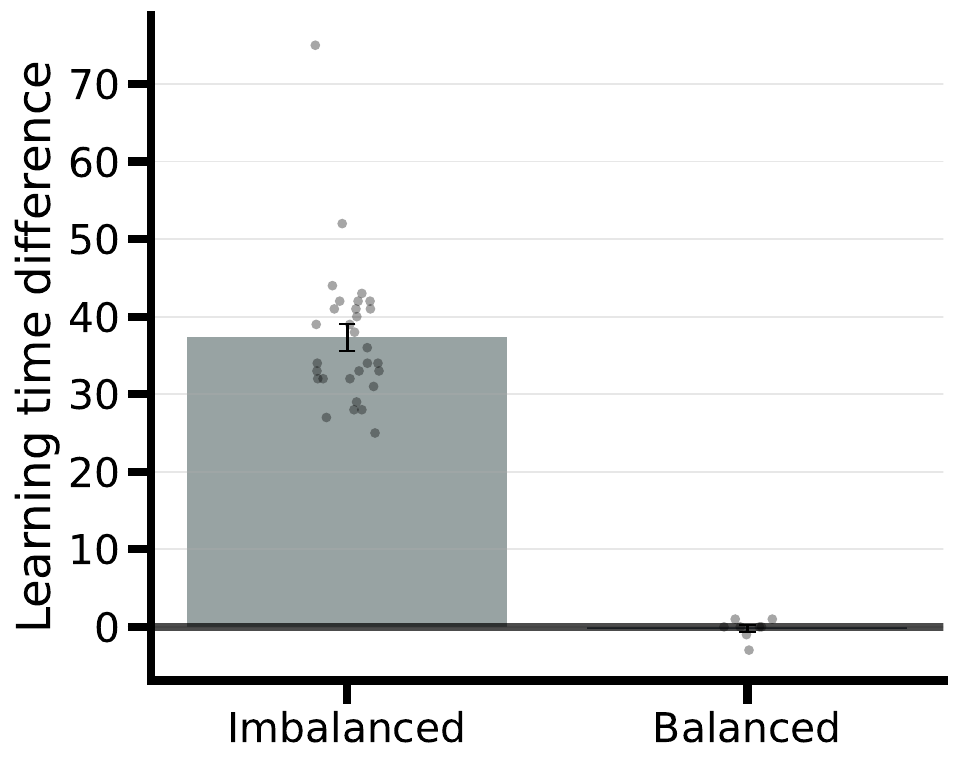}
\caption{Difference in learning time in epochs between the two sub-tasks, time of copy-first minus time of copy-last. For each seed, we computed the first epoch at which copy-first and copy-last each exceeded 90\% test accuracy, retaining only seeds that reached threshold on both tasks. Bars show the mean signed difference, with standard-error error bars and individual seeds overlaid.}
\label{fig:icl_learning_time_diff}
\end{figure}

\subsection{Development of model policy}

\begin{figure}[h]
\centering
\includegraphics[width=1\textwidth]{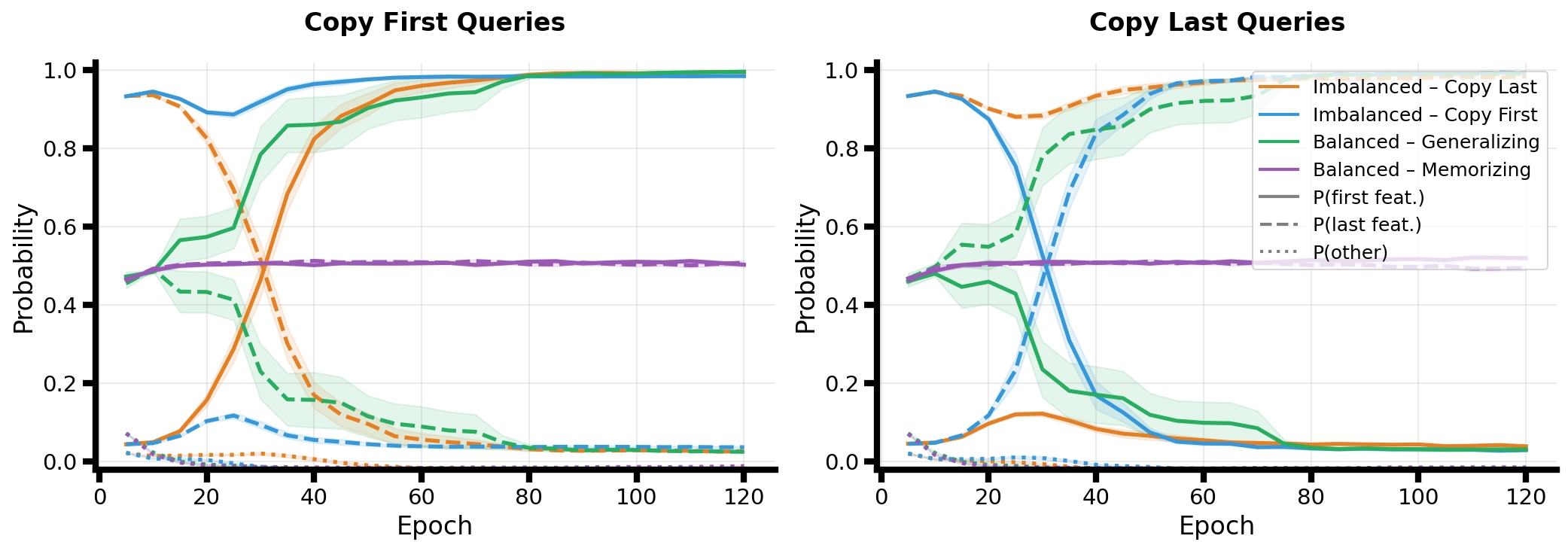}
\caption{\textbf{Probability distribution of the model over different groups of outputs (copy-first consistent, copy-last consistent, other).} Lines show the mean, shaded regions indicate $\pm$ s.e. (\textit{left}) Probability distribution outputted by different models over the course of training on held-out copy first queries. Models trained on the (light blue) imbalanced copy-first condition learn to quickly allocate almost all of their probability mass on (solid line) the first token of the sequence, and very little on (wide dashed line) the last token and even less on (thin dashed lines) all other tokens. Models trained on the (orange) imbalanced copy-last condition start out allocating most of their probability mass to the last token, even on this condition of copy first queries, but then learn to differentiate smoothly. Models trained in the interleaved condition evenly spread their probability mass between both the first and last tokens. They are divided into two groups, (green) generalizing models that slowly learn to accurately distinguish between first and last models and (purple) memorizing models that remain randomly guessing between the first and last solution over the whole of training. (\textit{right}) Same as left, but instead for copy last queries. Shaded region represents error of mean.}
\label{fig:icl-dynamics-behavior}
\end{figure}

\subsection{Linear probing for task identity}
\label{app:task_identity_decoding}

To quantify how readily the two subtasks are linearly represented in the residual stream after pretraining, we trained a logistic-regression probe on activations taken at the query position after each sublayer: the embedding output, after L1 attention, after L1 FFN, after L2 attention, and after L2 FFN. For each trained model, we constructed two single-task probe datasets, one from copy-first examples and one from copy-last examples, and evaluated probe accuracy using cross-validation (500 new task samples per task). Probe accuracy near $50\%$ indicates little linear separability, while higher values indicate that the two sub-tasks are linearly represented. We find that task identity is only strongly linearly represented in the later layers of models that are trained in the imbalanced curriculum and support our hypothesis around circuit entanglement in the balanced curriculum.

\begin{figure} [h]
\centering
\includegraphics[width=.6\textwidth]{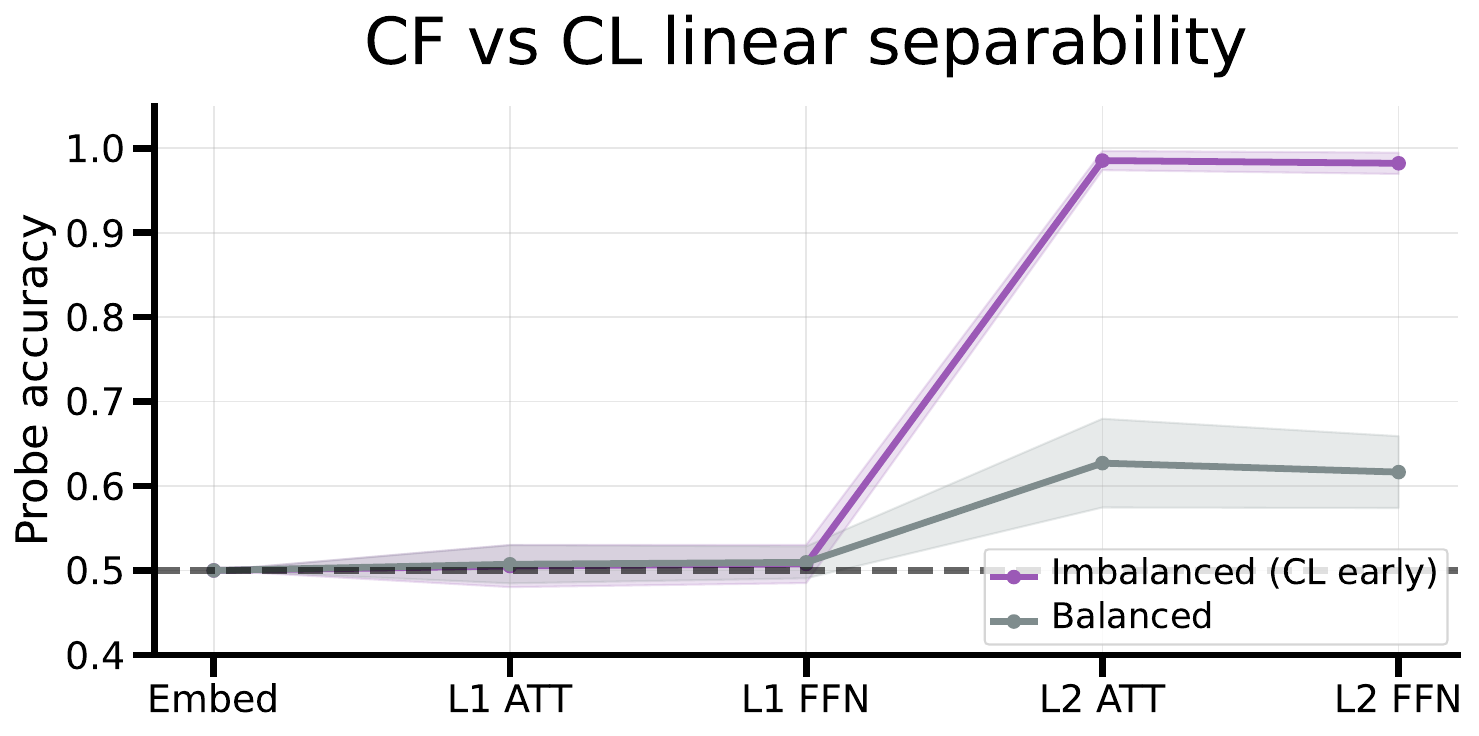}
\caption{\textbf{Task identity is linearly represented only for imbalanced models.} Linear probe accuracy for discriminating copy-first versus copy-last representations across sub-layers. Curves show mean probe accuracy across seeds for each training regime (shaded bands: $\pm$ s.d.). We find that only for models trained in the imbalanced curriculum task identity is strongly represented linearly after pretraining.}
\label{sfig:linear_probing_analysis}
\end{figure}

\subsection{Curriculum-dependent task entanglement can be measured directly}

\begin{figure} [h]
\centering
\includegraphics[width=0.65\textwidth]{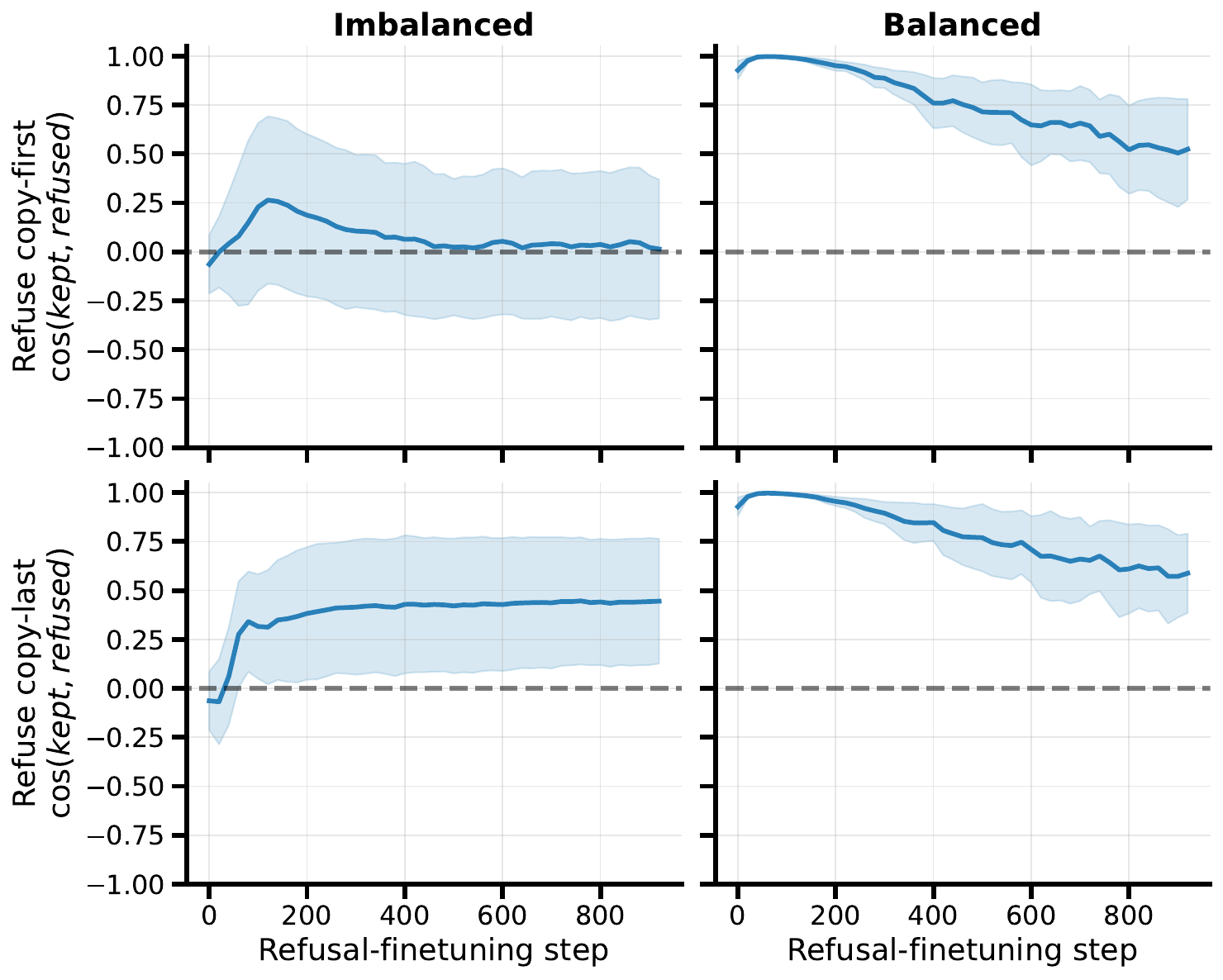}
\caption{\textbf{Residual stream task directions for the two tasks is closer to orthogonal under imbalanced curricula.}
    We show the cosine similarity between residual stream activity at the colon position at the final L2 FFN of the refused and the kept task during fine-tuning on paired prompts that only flip the task label. The columns split this comparison by the different curricula, rows by whether we refusal-train copy-first or copy-last. We focus on residual stream activity after the final FFN block (lines represent the mean across seeds, shaded regions indicate $\pm$s.d.). Under a balanced curriculum, the distance between task representations starts at close to one, maximal correlation, and then decreases during fine-tuning. For imbalanced curricula on the other hand, the distances start orthogonal, around zero. This provides some explanations as to why refusal fine-tuning more strongly affects the kept task after a balanced curriculum.}
\label{sfig:cos_dist_of_refuse_v_kept}
\end{figure}

Honing in on the previous result, we were interested in the similarity of residual stream activity for the two tasks across the refusal fine-tuning period. To this end, we computed the cosine similarity between these two task representations after the different curricula and while fine-tuning against different tasks, see \cref{sfig:cos_dist_of_refuse_v_kept}. Interestingly, this reveals that the curricula create vastly different starting configurations: Under the balanced curriculum, task representations are highly correlated at the outset, and only gradually decorrelate over the course of fine-tuning. Task representations under the imbalanced curriculum are nearly orthogonal (disentangled) at the start of refusal training, and then only slightly change in their cosine similarity during fine-tuning. This strongly suggests that the highly entangled representations under the balanced curriculum contribute to the observed overrefusal. Additionally, it suggests that monitoring the angle between related tasks during fine-tuning might serve as an early warning signal more generally, to detect whether fine-tuning has unwanted side effects on other capabilities.

\clearpage

\subsection{Inverse imbalance: Copy-first early}
\label{app:copy_first_early}

\begin{figure} [h]
\includegraphics[width=1\textwidth]{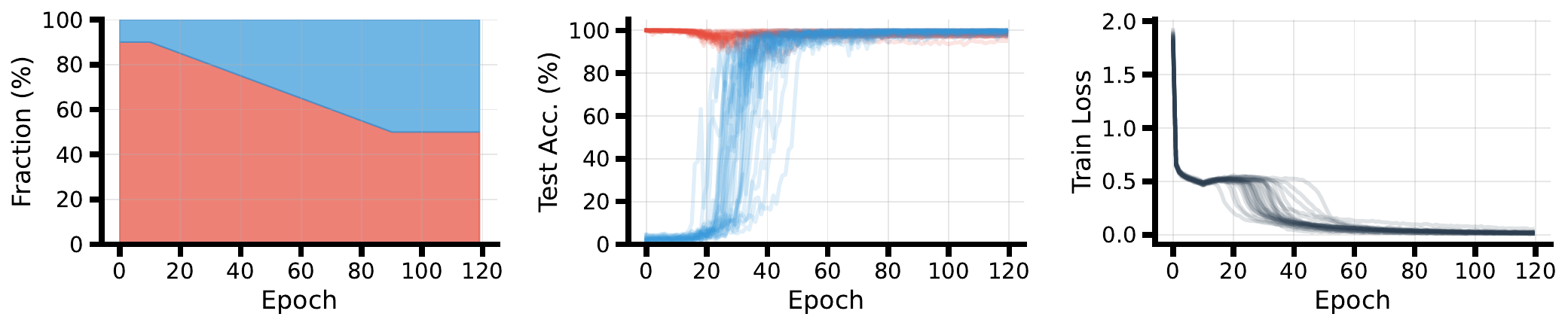}
\caption{\textbf{Training dynamics for the imbalanced copy-first-early curriculum.} Copy first is trained early, the task mixture is gradually annealed to 50:50. Left: task fraction over training. Center: held-out test accuracy on both tasks across seeds. Right: training loss. Under this curriculum, all $30/30$ seeds converge to generalizing solutions.}
\label{fig:ICL-copy-first_training}
\end{figure}

\paragraph{Acquisition dynamics.} For completeness, we also consider an imbalanced learning setting in which the copy-first task is introduced early in training. We find similar results for learning dynamics such that the imbalanced training helps models to consistently learn a generalizing solution \cref{fig:ICL-copy-first_training}. 

\begin{figure} [h]
\centering
\includegraphics[width=0.8\textwidth]{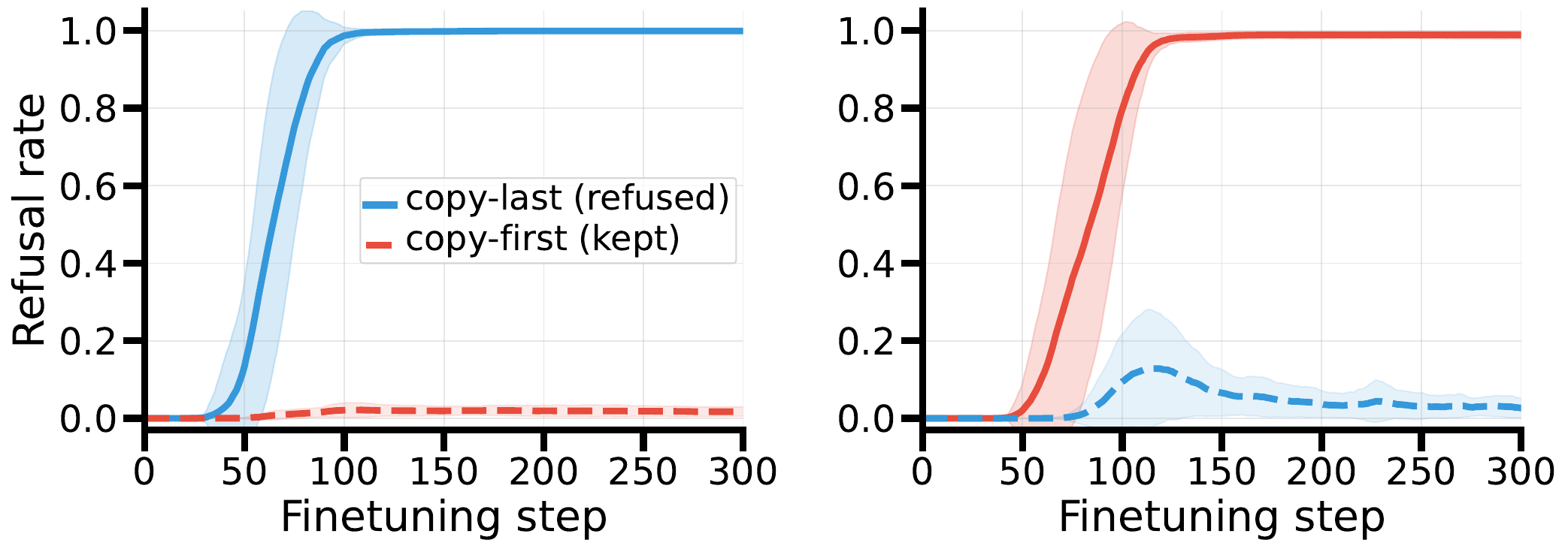}
\caption{\textbf{Refusal training introduces only minimal interference in the inverse imbalanced curriculum.} Same as \cref{fig:refusal_rate} but for the copy-first early imbalanced curriculum in \cref{fig:ICL-copy-first_training}, lines represent the mean with shaded regions indicating standard deviation over seeds. We find that models trained under this type of imbalanced curriculum similarly show less interference.}
\label{fig:ICL-copy-first-unlearning}
\end{figure}

\paragraph{Refusal results} We further find that models in this curriculum show similar effects during the refusal phase as seen in \cref{fig:ICL-copy-first-unlearning}. Models in this type of imbalanced curriculum, the refusal rate on the unrefused (kept) task similarly remains low throughout fine-tuning.

\clearpage
\subsection{Accuracy during refusal fine-tuning}

\label{app:accuracy-refusal-training}
\begin{figure} [h]
\centering
\includegraphics[width=\textwidth]{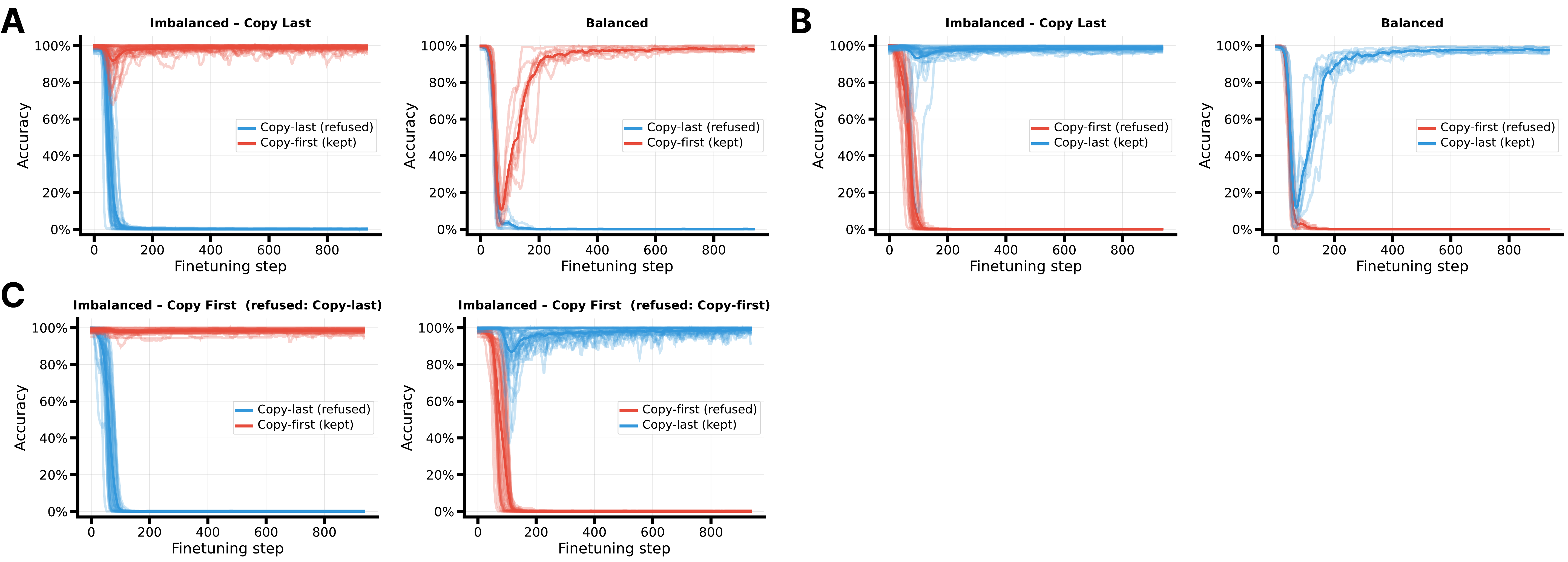}
\caption{\textbf{Accuracy on the kept task is maintained in imbalanced models over the course of refusal fine-tuning.} (\textbf{A}) Accuracy while refusal fine-tuning the copy-last task for both (\textit{left}) imbalanced curricula and (\textit{right}) balanced curricula from the main text. (\textbf{B}) Same as (A), but refusal fine-tuning on the copy-first task. (\textbf{C}) Results for the imbalanced curriculum in which the copy-first task is introduced early in training while refusal fine-tuning on (\textit{left}) copy-first and (\textit{right}) copy-last.}
\label{sfig:accuracy-refusal-training}
\end{figure}

Beyond refusal rates we also examine accuracy on the kept (non-refused) task throughout fine-tuning. We find that accuracy on the held-out tasks does not decrease significantly during refusal fine-tuning for the imbalanced curricula, and is degraded but then recovers for the held-out task (\cref{sfig:accuracy-refusal-training}). These results exactly mirror observations in \cref{fig:refusal_rate} but show that models trained under imbalanced curricula do not only not over-refuse but also retain performance on the original task, while performance in balanced models temporarily degrades due to overrefusals.

\subsection{Constant imbalanced training data distributions}
\label{app:constant_imbalanced_icl}

\begin{figure} [h]
\centering
\includegraphics[width=.8\textwidth]{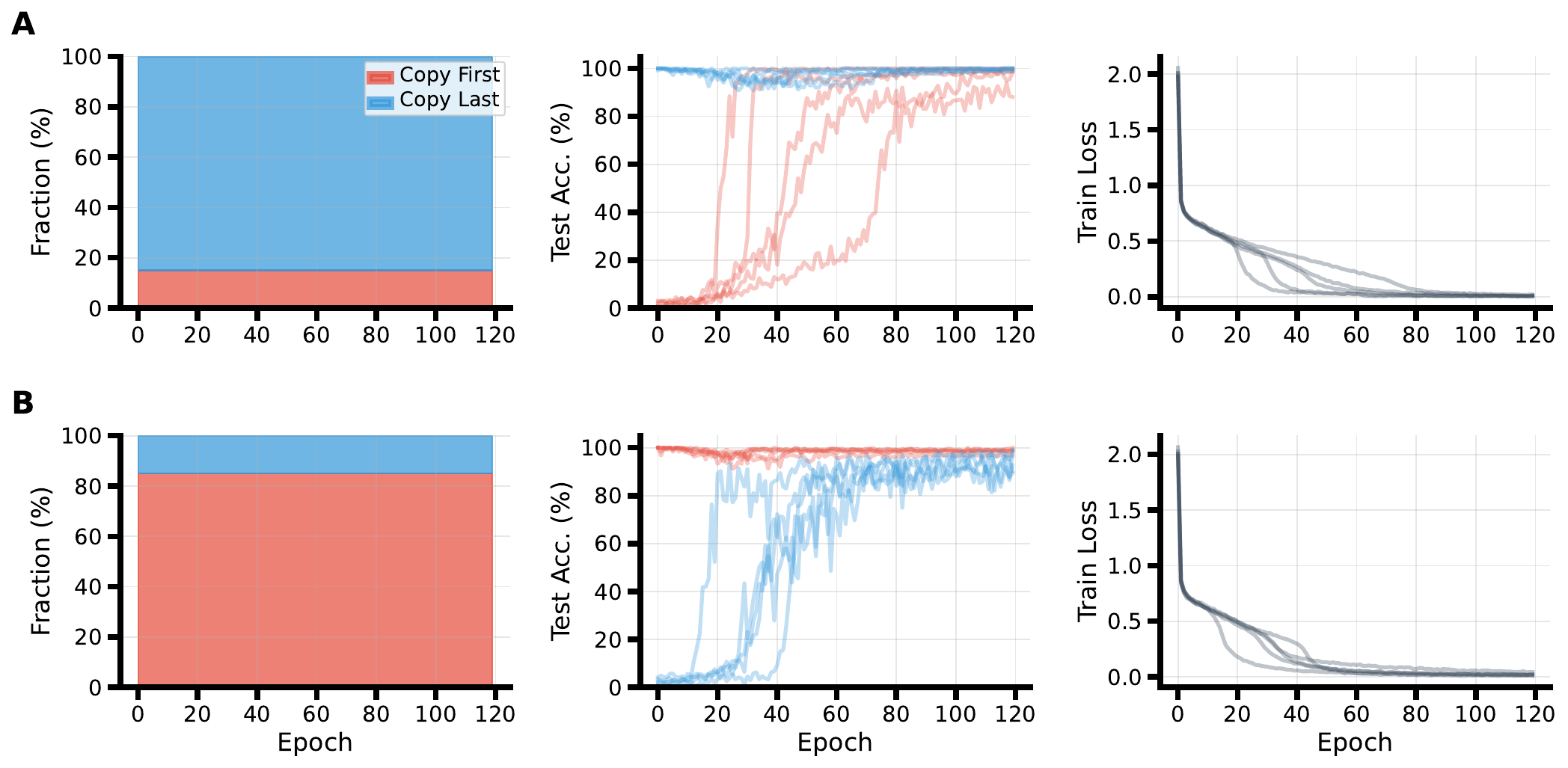}
\caption{\textbf{Training dynamics for constant imbalanced datasets.} Imbalanced training tasks without linear fading also allow models to learn generalizing solutions albeit at a slightly slower rate (compare to \cref{fig:task_setup_and_training}) Left: task fraction over training. Center: held-out test accuracy on both tasks across seeds. Right: training loss. Under this curriculum, all $5/5$ seeds converge to generalizing solutions.}
\label{fig:ICL-constant curricula}
\end{figure}

\begin{figure} [h]
\includegraphics[width=1\textwidth]{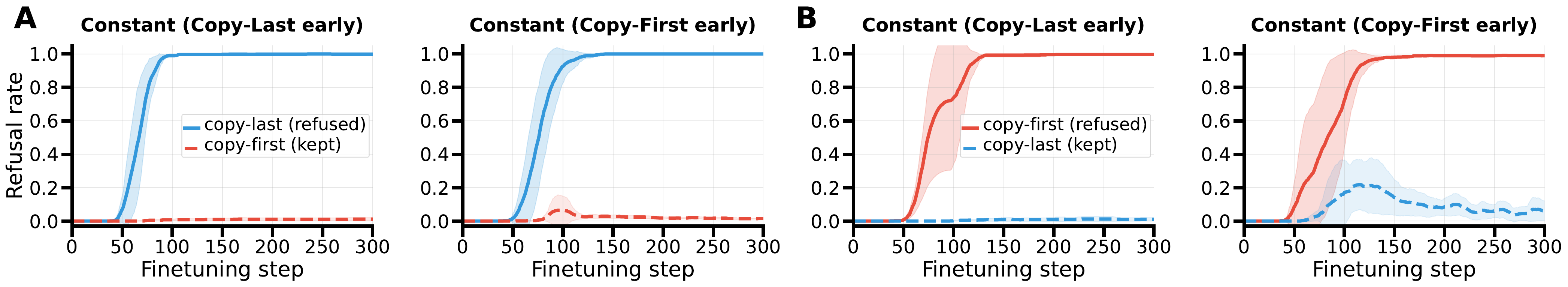}
\caption{\textbf{Refusal training introduces only minimal interference in the constant imbalanced regime.} Same as \cref{fig:refusal_rate} but for constant imbalanced task mixtures shown in \cref{fig:ICL-constant curricula}, lines represent the mean with shaded regions indicating standard deviation over seeds. We find that models trained under this type of imbalanced curriculum similarly show reduced interference when selectively unlearning one of the two sub-tasks.}
\label{fig:ICL-constant_curricula_refusal}
\end{figure}

\paragraph{Acquisition dynamics} We further consider imbalanced training data distributions that do not contain explicit linear fading of the task mixture but manipulate the order of task acquisition via two fixed constant task mixtures of 15:85 and 85:15 (copy first:copy last). We train 5 seeds per condition with all other hyperparameters as in \cref{sec:icl_dataset} and \cref{app:icl_task}. We find that this manipulation similarly encourages the learning of generalizing solutions as seen in \cref{fig:ICL-constant curricula} such that all $5/5$ seeds learn to generalize (>90 accuracy). These results imply that task imbalance beyond fading can also encourage generalization. However, we note that seeds in this regime converge slower than in \cref{fig:task_setup_and_training}. These results should be seen as a form of implicit curriculum where the data distribution indirectly encourages the sequential learning of both sub-tasks. It is also noteworthy that these results rule out that the dynamic resampling of tasks seen in the imbalanced curricula is the driver of improved generalization behavior. We further find that models in this regime also show reduced overrefusals during fine-tuning (see \cref{fig:ICL-constant_curricula_refusal}).

\clearpage

\clearpage

\subsection{Training with infinite data}
\label{app:infinite data regime}
The results in the main paper used a fixed training set of $S = 20{,}000$ sequences that the model revisits every epoch. However, given the importance of data distributional properties for the emergence of ICL \citep{chan_data_2022, reddy_mechanistic_2023, singh2023TransientNatureEmergent} we also evaluate models trained in a regime of \emph{infinite} data were we sample new sequences every epoch. Concretely, at the start of each epoch $t$ the training set is re-sampled, yielding a fresh draw of $S = 20{,}000$ sequences. All other Hyperparameters are identical to the fixed-data runs. This increases the effective diversity of training samples, ensuring the model cannot overfit to a fixed pool of samples and instead must learn the underlying task structure directly.

\begin{figure} [h]
\centering
\includegraphics[width=1\textwidth]{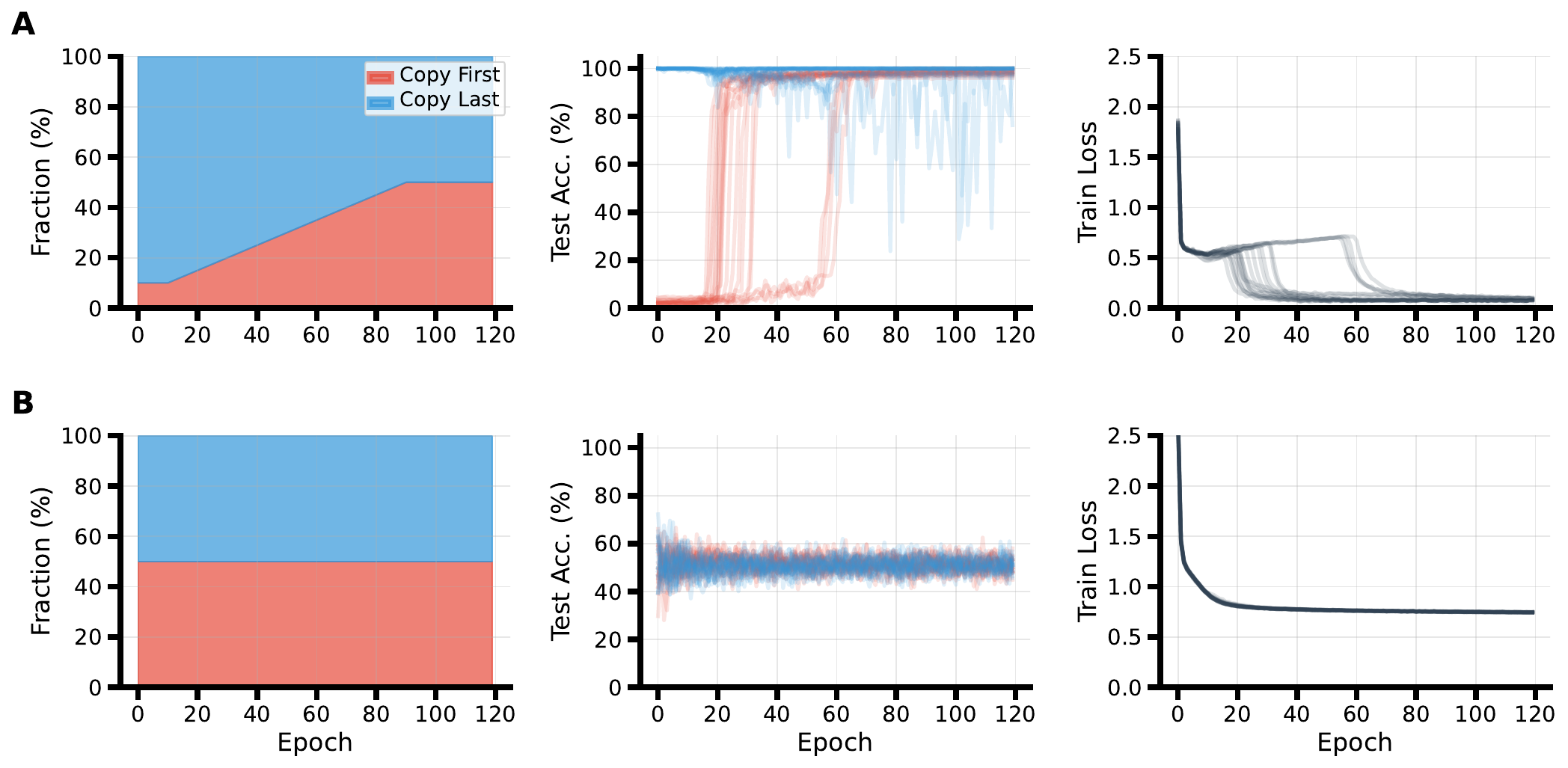}
\caption{\textbf{Training curricula in the infinite data regime.} Same as \cref{fig:task_setup_and_training} but for infinite data.}
\label{fig:ICL_infinite_data}
\end{figure}

\Cref{fig:ICL_infinite_data}B shows that surprisingly, models trained under the balanced curriculum do not learn the generalizing solution such that $0/20$ seeds converge to a generalizing solution within 120 epochs. While for the imbalanced regime (\cref{fig:ICL_infinite_data}A) $19/20$ seeds acquire a generalizing ICL solution. These results underscore that imbalanced curricula also reliably produce generalizing solutions in the infinite data regime. However, it is perhaps surprising that in our multi task setting greater diversity reduced the propensity of models to successfully acquire ICL solutions in the balanced curriculum. Previous results have demonstrated that greater data diversity typically facilitates the acquisition of ICL solutions \citep{chan_data_2022}. In contrast, the result in this section shows that for task recognition modes of ICL \citep{pan2023WhatInContextLearning} perhaps the partial acquisition of memorizing, in-weights solutions facilitates in-context learning in a more cooperative manner \citep{singh2025StrategyCoopetitionExplains}.

\clearpage
\section{Interpretability analyses: the controlled in-context learning task}
\label{app:mechanism_icl}

Both analyses are performed on each pretrained checkpoint and evaluated on the held-out $500$-sequence 50:50 test set described in \cref{app:icl_task}. As elsewhere in the paper, analyses are restricted to seeds that reach $\geq 90\%$ held-out accuracy on both tasks at the end of pretraining; results are reported per task (copy-first vs copy-last) and per curriculum.

\subsection{Mean-component ablation}
\label{app:ablation_icl}

For each candidate sublayer (L1 ATT, L1 FFN, L2 ATT, L2 FFN), we first compute the position-wise dataset-mean activation of that sublayer's output over the evaluation set. We then re-run the model with that sublayer's output replaced by its mean (a mean ablation that preserves the global activation statistics but
removes input-conditional information, \citep{wang2022InterpretabilityWildCircuit}), and measure per-task accuracy
on the same held-out set. The unablated baseline is included for comparison. Reported figures focus on attention block ablations.

\subsection{Direct logit attribution}
\label{app:dla_icl}
Direct logit attribution (DLA) decomposes output logits into the direct contributions of components that write to the residual stream \citep{elhage2021mathematical,wang2022InterpretabilityWildCircuit}. We apply DLA at the query colon position. Let $h$ denote the final pre-LayerNorm residual stream at this position, decomposed as $h=\sum_i c_i$, where each $c_i$ is the embedding contribution or the output of an attention/FFN sublayer. Since the final unembedding is applied after the final LayerNorm, we include the learned LayerNorm scale $\gamma$ and the per-token residual-stream standard deviation $\sigma(h)$. Treating $\sigma(h)$ as fixed at its forward-pass value, the direct contribution of component $c_i$ to the correct-token logit $y$ is
\[
    \mathrm{DLA}_i(y)
    =
    \left[
    (c_i/\sigma(h)) (W_U \gamma)^\top
    \right]_y .
\]
This fixed-scale approximation makes the final LayerNorm linear in the residual-stream components, so component contributions are additive and approximately reconstruct the correct-token logit. We report per-component contributions on the held-out set, separately for copy-first and copy-last queries.

\subsection{Activation patching}
\label{app:patching_icl}
Activation patching is a technique that carefully replaces specific internal activation of transformers during a forward pass from a previous run \citep{heimersheim2024HowUseInterpret, meng2023LocatingEditingFactual}. In this paper we use this technique to causally reveal which model components are implicated in the solution of the learned copy-first and copy last sub-tasks. Here we specifically use \emph{denoising} activation patching to localize the pathways through which each sub-task is implemented. 

\paragraph{Corruption.}
For each "clean" sequence drawn from one task, we construct a "corrupt" counterpart by recomputing the label of every in-context example under the \emph{other} task's rule; all features (including the query feature) are left unchanged. Under the clean input the model follows the in-context rule and predicts the correct label. Under the corrupt input it follows the flipped rule and predicts the other task's answer, dropping accuracy on the original target to near chance.

\paragraph{What is patched.}
A denoising patch runs the forward pass on the corrupt (label flipped) input but overwrites one or more activations with the corresponding values from the clean forward pass; every other activation is recomputed from the corrupted input. For L2 attention we patch the sublayer's
Q-, K-, and V-projections individually and pairwise and triple combinations (QK, KV, QV and QKV restoration). For L2 FFN we patch the residual read (the residual stream immediately before the FFN sublayer). While the main paper only reports patching results from L2 Q, K, V, and FFN as they provide the decisive contrast, we conduct and report the patching of combinations for completeness.

\paragraph{Recovery metric.}
For each patch we compute recovery as
\[
\;
\frac{\mathrm{Acc}_{\mathrm{patched}} - \mathrm{Acc}_{\mathrm{corrupt}}}
     {\mathrm{Acc}_{\mathrm{clean}} - \mathrm{Acc}_{\mathrm{corrupt}}},
\]
separately for copy-first and copy-last queries. A recovery of $0$ corresponds to no improvement over the corrupt baseline; $1$ corresponds to full restoration of clean-run accuracy.

\paragraph{Full table.}
Table~\ref{tab:patch_full} reports every patch we evaluated under both training conditions. ``Clean'' and ``Corrupt'' rows give raw accuracy. Rows give the accuracy and recovery fraction obtained when a given channel is patched; Q,K,V-combinations (e.g.\ ``Restore QK'') patch multiple projection inputs jointly.

\begin{table}[t]
\centering
\footnotesize
\setlength{\tabcolsep}{4pt}
\renewcommand{\arraystretch}{1.08}

\textbf{Imbalanced}\\[3pt]
\begin{tabular}{lcccc}
\toprule
Patch & CF Acc & CL Acc & CF Rec. & CL Rec. \\
\midrule
Clean
  & $99.6 \pm 0.5$
  & $98.4 \pm 1.1$
  & ---
  & --- \\
Corrupt
  & $3.8 \pm 1.6$
  & $2.8 \pm 1.2$
  & (base)
  & (base) \\
L2 Q
  & $3.8 \pm 1.6$
  & $2.8 \pm 1.2$
  & $0.0 \pm 0.0$
  & $0.0 \pm 0.0$ \\
L2 K
  & $81.6 \pm 20.0$
  & $39.8 \pm 32.7$
  & $81.2 \pm 20.7$
  & $39.0 \pm 34.4$ \\
L2 V
  & $22.5 \pm 21.7$
  & $21.2 \pm 20.9$
  & $19.7 \pm 22.8$
  & $19.3 \pm 21.6$ \\
L2 QK
  & $81.6 \pm 20.0$
  & $39.8 \pm 32.7$
  & $81.2 \pm 20.7$
  & $39.0 \pm 34.4$ \\
L2 KV
  & $99.6 \pm 0.5$
  & $98.4 \pm 1.1$
  & $100.0 \pm 0.0$
  & $100.0 \pm 0.0$ \\
L2 QV
  & $22.5 \pm 21.7$
  & $21.2 \pm 20.9$
  & $19.7 \pm 22.8$
  & $19.3 \pm 21.6$ \\
L2 QKV
  & $99.6 \pm 0.5$
  & $98.4 \pm 1.1$
  & $100.0 \pm 0.0$
  & $100.0 \pm 0.0$ \\
L2 FFN
  & $56.2 \pm 35.5$
  & $88.1 \pm 22.7$
  & $54.9 \pm 37.1$
  & $89.3 \pm 23.8$ \\
\bottomrule
\end{tabular}

\vspace{0.8em}

\textbf{Balanced}\\[3pt]
\begin{tabular}{lcccc}
\toprule
Patch & CF Acc & CL Acc & CF Rec. & CL Rec. \\
\midrule
Clean
  & $99.6 \pm 0.4$
  & $99.2 \pm 0.7$
  & ---
  & --- \\
Corrupt
  & $2.7 \pm 0.7$
  & $2.5 \pm 1.2$
  & (base)
  & (base) \\
L2 Q
  & $2.7 \pm 0.7$
  & $2.5 \pm 1.2$
  & $0.0 \pm 0.0$
  & $0.0 \pm 0.0$ \\
L2 K
  & $99.4 \pm 0.4$
  & $98.4 \pm 0.9$
  & $99.8 \pm 0.3$
  & $99.3 \pm 0.6$ \\
L2 V
  & $3.4 \pm 1.0$
  & $3.0 \pm 1.3$
  & $0.7 \pm 0.5$
  & $0.5 \pm 0.6$ \\
L2 QK
  & $99.4 \pm 0.4$
  & $98.4 \pm 0.9$
  & $99.8 \pm 0.3$
  & $99.3 \pm 0.6$ \\
L2 KV
  & $99.6 \pm 0.4$
  & $99.2 \pm 0.7$
  & $100.0 \pm 0.0$
  & $100.0 \pm 0.0$ \\
L2 QV
  & $3.4 \pm 1.0$
  & $3.0 \pm 1.3$
  & $0.7 \pm 0.5$
  & $0.5 \pm 0.6$ \\
L2 QKV
  & $99.6 \pm 0.4$
  & $99.2 \pm 0.7$
  & $100.0 \pm 0.0$
  & $100.0 \pm 0.0$ \\
L2 FFN& $2.9 \pm 0.7$
  & $2.6 \pm 1.2$
  & $0.2 \pm 0.2$
  & $0.1 \pm 0.2$ \\
\bottomrule
\end{tabular}
\vspace{7pt}
\caption{Denoising activation patching across all evaluated channels.
Accuracies and recovery fractions are reported in percent as mean $\pm$
standard deviation across seeds.}
\label{tab:patch_full}
\end{table}

\paragraph{Why does restoring L2 KV recover copy-last under Imbalanced
training, when copy-last has been argued to bypass Block~2 attention?} For imbalanced training, mean ablation (\cref{sec:ablation_results_icl}) shows that \emph{removing} L2 attention leaves copy-last largely intact, so the sublayer is not \emph{necessary} for copy-last. Denoising patching with restored \cref{tab:patch_full} KV shows that supplying clean K and V to L2 attention \emph{recovers} copy-last to clean-run accuracy, so the sublayer is \emph{sufficient} to perform copy-last. These two statements are not in conflict: L2 attention has the capacity to solve copy-last but does not need to, because L1 attention already writes the copy-last answer into the residual stream that L2 FFN reads. The two-pathway claim is therefore about \emph{primary routing}, not about hard capability boundaries: copy-first is primarily routed through K (L2~ATT) and copy-last through the FFN residual read, but
both sublayers retain residual capacity to solve either task. The Balanced condition lacks this redundancy: restoring the FFN residual read alone yields essentially zero recovery on either task, so all routing flows through K.

\clearpage
\section{Mechanistic analyses on refusal fine-tuned model}
\label{app:refusal_mechanism}
\begin{figure}[h]
\centering
\includegraphics[width=1\textwidth]{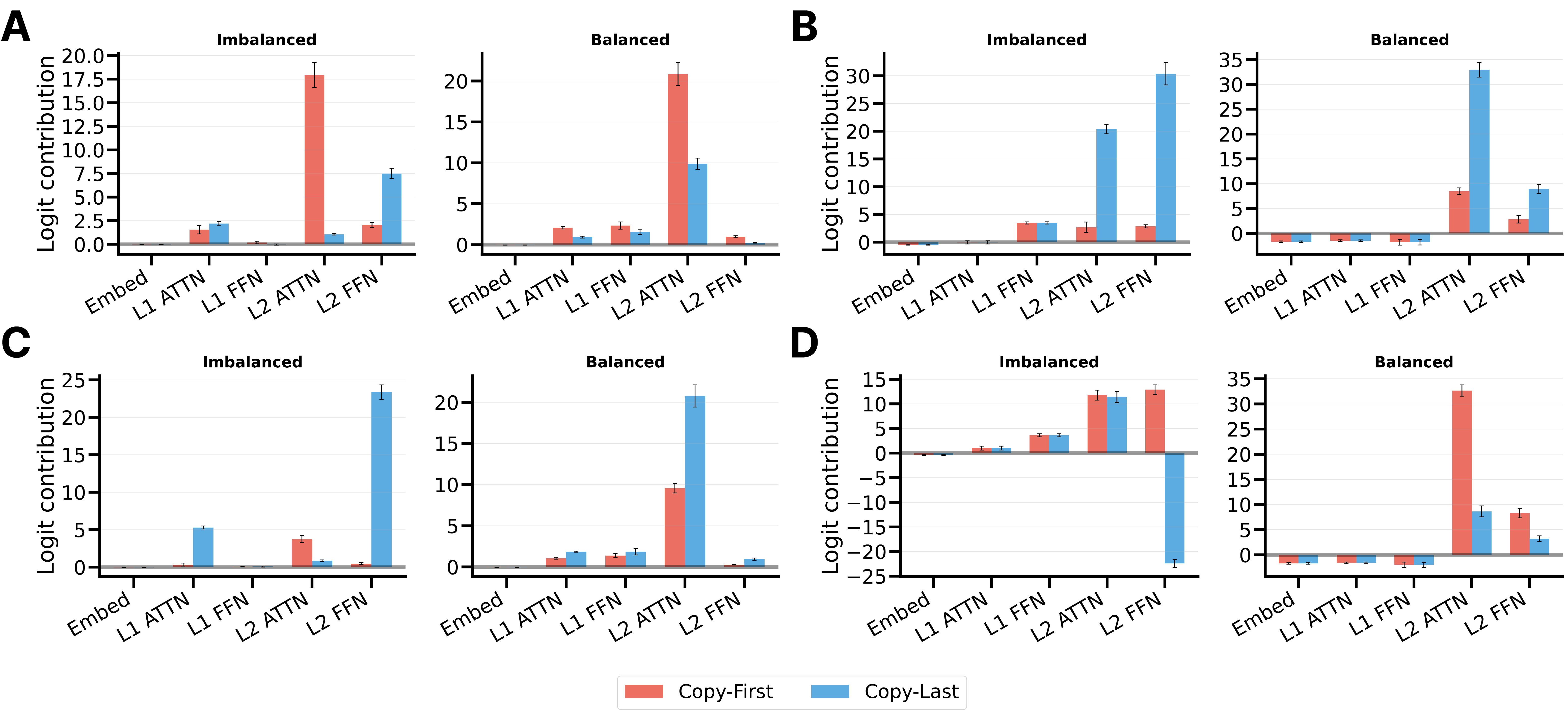}
\caption{\textbf{DLA analysis on refusal fine-tuned models reveals that refusal is primarily implemented in the last two layers.} (\textbf{A}) Logit contributions for the correct token under the original task of different components for models fine-tuned to refuse the copy-last task. Models were pretrained under curricula that were (left) balanced and (right) imbalanced. (\textbf{B}) Same as (A), but on the refusal token. (\textbf{C}) Same as (A), but for models fine-tuned to refuse the copy-first task. (\textbf{D}) Same as (C), but for the refusal token. Error bars depict the standard error of the mean.}
\label{fig:dla-refusal-balanced}
\end{figure}

\clearpage
\section{Experimental setup for the SLL task}
\label{app:sll_setup}

\subsection{Synthetic language grammar}

We construct a synthetic language of profile-style sentences from a fixed grammar over features. Each profile sentence begins with ``This person'' followed by a number of feature descriptions, for example: ``This person is named Bob, lives in Paris, speaks French, has a pet lizard, and drinks tea''. 

For a given feature, a value is selected from a finite pool (e.g. NAME $ \in \{ \text{Alice, Bob, Charlie, Diana} \}$ ). Each feature value is paired with a preamble (e.g. ``is named Bob'', ``is called Bob''). A feature value can be combined with any corresponding feature preamble, and the features can appear in any arbitrary order in the sentence.

In this experiment, the grammar consists of 9 features, and each feature was associated with a finite pool of 4 values and 2 preambles. The sentences have a fixed length of 5 features, such that the general form is:

\begin{center}
    This person $<\text{Preamble}_{1}>$ $<\text{Value}_{1}>$ ... $<\text{Preamble}_{5}>$ $<\text{Value}_{5}>$
\end{center}

We used a discrete vocabulary in which each preamble (e.g. ``is named'') and feature value (e.g. ``Bob'') corresponds to a unique token. We do not replicate the full system here, since any set of unique tokens can be used to represent the feature values.

\subsection{Rule system}
\label{app:sll_rulesystem}

On top of the grammar, we construct a system of rules to operationalize the concept of alignment for our task. Rules take the form of specifying a condition (e.g. "NAME=Bob") and a target (e.g. "JOB=Builder"). If a sentence meets the condition of a rule, it also needs to meet the target, otherwise the sentence is considered misaligned. If a sentence does not specify the targeted feature then the sentence counts as aligned by default. A sentence is considered aligned only if it satisfies all applicable rules. If two rules clash by specifying different target values for the same feature, the higher priority rule takes precedence. 

\begin{table}[h]
\centering
\begin{tabular}{llll}
\toprule
Rule & Condition & Target & Priority \\
\midrule
$R_1$ & NAME $=$ Alice & JOB $=$ Builder & 10 \\
$R_2$ & SPORT $=$ Tennis & LOCATION $=$ Sydney & 5 \\
$R_3$ & PET $=$ Rat & MUSIC $=$ Rock & 7 \\
$R_4$ & MUSIC $=$ Pop & LOCATION $=$ Chicago & 15 \\
$R_5$ & PET $=$ Crane & JOB $=$ Doctor & 18 \\
\bottomrule
\end{tabular}
\vspace{7pt}
\caption{\textbf{SLL task rule system}. We employed 5 rules, two pairs of which had the potential to clash ($R_1$ and $R_5$ on JOB, $R_2$ and $R_4$ on LOCATION.)}
\end{table}

The grammar and rule system are independent, a sentence can therefore be either: ungrammatical, if it violates the grammar, grammatical but misaligned, if it follows the grammar but not the rule system, or grammatical and aligned, if it follows both. Additionally, we draw a distinction between explicitly aligned sentences (those that trigger at least one rule and satisfy all triggered rules) and neutral sentences (those that are aligned by default as they do not trigger any rules). 

\subsection{Dataset construction and splits}
\label{app:sll_aligned_data}

We randomly sample $400$ neutral, $400$ explicitly aligned (rule condition and target met), and $400$ misaligned sentences for training. 

In order to probe whether the model has learned each rule and can generalize to unseen prompts, we construct a test set of single rule elicitation prompts, each triggering solely one rule. We sample 50 elicitation prompts for each rule (250 total). To ensure that none of these prompts appear in the training set, we exclude any sentences containing the elicitation prompts from the training set.

\subsection{Curricula}
\label{app:sll_curricula}

During pretraining, models are trained on a mixture of aligned and misaligned sentences. At each epoch $t$, a sampling distribution over these datasets is defined by the weight assigned to misaligned examples, $w_{\text{misaligned}}(t)$. The remaining probability mass is split evenly between explicitly aligned and neutral data. 
\begin{equation}
w_{\text{aligned}}(t) = w_{\text{neutral}}(t) = \frac{1 - w_{\text{misaligned}}(t)}{2}
\end{equation}

We consider two curriculum conditions:

\paragraph{Balanced}
The proportion of misaligned data is held constant throughout pretraining:
\begin{equation}
w_{\text{misaligned}}(t) = \rho,
\end{equation}
where $\rho$ is the corruption level (here $\rho = 0.4$).

\paragraph{Imbalanced}
The proportion of misaligned data increases over training. We use a piecewise schedule with an initial plateau followed by a linear ramp:
\begin{equation}
w_{\text{misaligned}}(t) =
\begin{cases}
w_{\text{low}}, & t < T_{\text{plateau}} \\
w_{\text{low}} + (w_{\text{high}} - w_{\text{low}})\cdot \tau(t), & t \geq T_{\text{plateau}},
\end{cases}
\end{equation}
where $\tau(t)$ linearly interpolates from $0$ to $1$ over the remaining epochs.

To ensure curricula are matched for the total exposure to misaligned data, we choose $w_{\text{high}}$ such that the average proportion of misaligned data matches the target corruption level:
\begin{equation}
\frac{1}{T} \sum_{t=1}^{T} w_{\text{misaligned}}(t) = \rho.
\end{equation}

We train for $800$ epochs on the pretraining distribution, followed by $200$ fine-tuning epochs on a distribution with no misaligned examples, a 50:50 mixture of explicitly aligned and neutral data.

\subsection{Architecture and optimization}
\label{app:dlr_architecture}

We use a pre-norm $4$-layer Transformer with $d_{\text{model}} = 128$,
$4$ heads per layer (head dim $32$), a GELU FFN with hidden dimension
$4\,d_{\text{model}} = 512$, rotary positional embeddings, and dropout $0.1$.
Optimization uses Adam with learning rate $5\times10^{-4}$, batch size $100$,
and gradient-norm clipping at $1.0$. We train $20$ random seeds and save
checkpoints every $40$ epochs during pretraining. All models were trained in PyTorch.

\subsection{Compute}

We used NVIDIA GeForce RTX 3090 GPUs for this study. For the main SLL experiments, training a single seed and a single curriculum took just under three minutes of GPU time, leading to a total compute time of two GPU hours (20 seeds times 2 curricula in the main text).

\clearpage

\section{Additional results for the SLL task}
\label{app:sll_extra}

\subsection{Additional control curricula}
\label{app:sll_extra_curricula}

\begin{figure}[htbp!]
\centering
\includegraphics[width=1\textwidth]{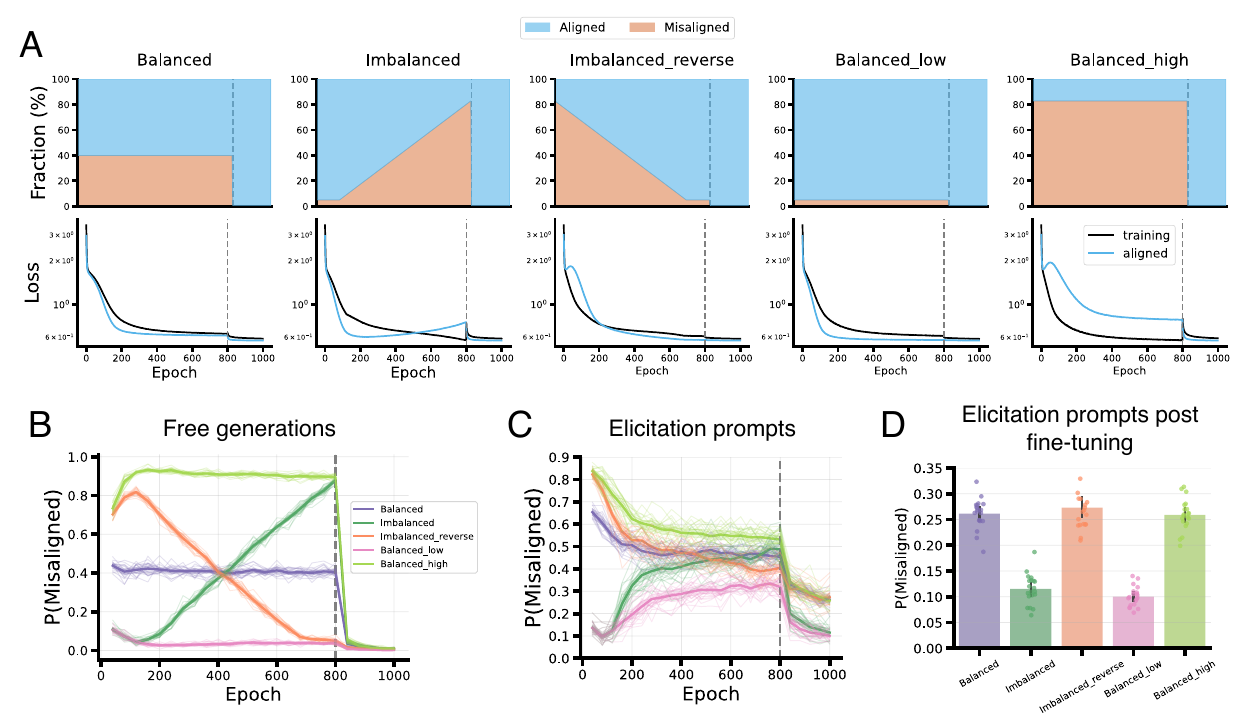}
\caption{\textbf{Additional curricula for SSL task}. 
        (\textbf{A}) Top row: To properly control our curriculum manipulation, we train models under additional regimes, beyond the balanced and imbalanced one presented in the paper. Critically, the imbalanced reverse curriculum starts with a large proportion of misaligned data, which decreases over time, flipping the order of the imbalanced curriculum.
        Bottom: Loss on the full training data (dark blue) and the aligned subset (light blue), shaded regions indicating $95$\% CI are included but not visible. 
        (\textbf{B}) Probability of misaligned sentences during free generations under different curricula. This probability tracks the epoch-to-epoch prevalence of misaligned data under the curriculum relatively closely (thin: individual seeds; bold: mean; dashed line: start of fine-tuning). (\textbf{C}) Elicitation prompts, however, reveal different degrees of robustness of rule-adherence (same plotting convention as C). (\textbf{D}) Mean probability of a misaligned completion for elicitation prompts under different curricula. After fine-tuning, the imbalanced curriculum performs on par with a curriculum which encountered a low level of misaligned data throughout the entirety of training (balanced low). On the other hand, the reverse imbalanced curriculum performs as poorly as balanced high, which experienced a constantly high proportion of misaligned data (error bars indicate $95$\% CI over seeds; individual seeds: dots).}
\label{fig:sll-control-curricula}
\end{figure}

Previously we compared two curricula (balanced and imbalanced) matched in total exposure to misaligned data. We now compare these to three additional controls, (\cref{fig:sll-control-curricula}A):

\begin{itemize}
    \item \textbf{Imbalanced\_reverse}: The proportion of misaligned data decreases from 80\% to 5\% across pretraining, i.e. the reverse of the imbalanced curriculum. 
    \item \textbf{Balanced\_low}: A constant low proportion of misaligned data (5\%)
    \item \textbf{Balanced\_high}: A constant high proportion of misaligned data (80\%)
\end{itemize}

We first examine $\text{P}(\text{Misaligned})$ of freely generated sentences. This closely tracks the training distribution across curricula, and fine-tuning results effectively suppresses misaligned generations in all cases with negligible differences between them, \cref{fig:sll-control-curricula}B. 

We next analyze $\text{P}(\text{Misaligned})$ for elicitation prompts during pretraining. Balanced\_high and balanced\_low converge to high and low plateaus respectively, while imbalanced\_reverse decreases gradually, mirroring the reversal for the imbalanced curriculum. These trends qualitatively reflect the training distribution. 

Fine-tuning leads to a reduction in $\text{P}(\text{Misaligned})$ for all curricula, however, the magnitude of improvement varies. Imbalanced and balanced\_low achieve similarly low levels of misalignment, suggesting limited exposure to misalignment early in training is as effective as maintaining low exposure throughout training. Furthermore, imbalanced\_reverse closely matches balanced\_high after fine-tuning, indicating that the impact of early exposure to high levels of misaligned data is not mitigated by cleaner data later in training. Interestingly, imbalanced\_reverse performs better than imbalanced at the end of pretraining, yet fine-tuning is less effective at reducing misalignment. Together these results highlight that early training has lasting effects on alignment that cannot be easily reversed by later training or fine-tuning.

\subsection{Grammaticality of free generations and elicitations}
\label{app:sll_beh_grammar}

\begin{figure}[htbp!]
\centering
\includegraphics[width=1\textwidth]{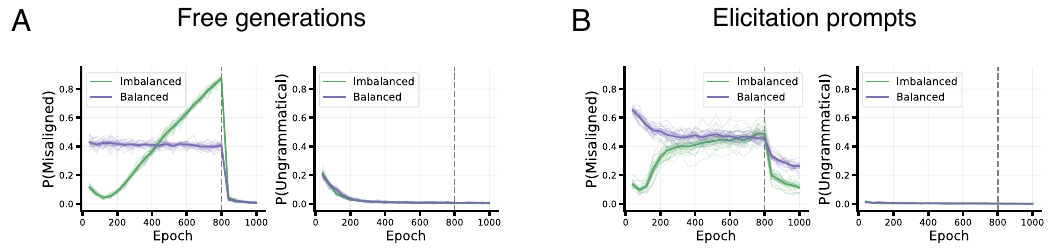}
\caption{\textbf{Misalignment and grammaticality of free generations and elicitations.}
         $\text{P}(\text{Misaligned})$ (left) and $\text{P}(\text{Ungrammatical})$ (right) for free generations (\textbf{A}) and elicitation prompts (\textbf{B}) across training under the imbalanced (green) and balanced (purple) curricula (thin: individual seeds; bold: mean; dashed line: start of fine-tuning).}
\label{fig:sll-beh-grammar}
\end{figure}

\subsection{Elicitation prompt alignment with rule system complexity and data set size}
\label{app:sll_rulesys_robust}

To establish the robustness of our finding that fine-tuning was improved after imbalanced training, we varied the complexity of the rule system and data set size, extending beyond the fixed training set used for the main text. We left all attributes of the rule system and training process untouched, except:
\begin{enumerate}
    \item We increased the number of values (Alice, Bob, ...) for each feature (NAME, JOB, ...) from 4 to 15, to allow for more rule variety.
    \item We vary the number of rules (from 5 to 40), and randomly sample that number of rules for each run independently.
    \item We vary the number of sentences making up the data set, while keeping the number of sentences presented during each "epoch" the same (1200, which is the total data set size after combining 400 neutral, aligned, and misaligned sentences in the original data set).
\end{enumerate}
When sampling rules, we ensure that each condition (e.g. NAME=Alice) occurs in at most one rule, but do not control the rule target, other than ensuring that it targets a different feature than the condition (as a result, multiple rules may demand the same specific target feature).

\begin{figure}[htbp!]
\centering
\includegraphics[width=1\textwidth]{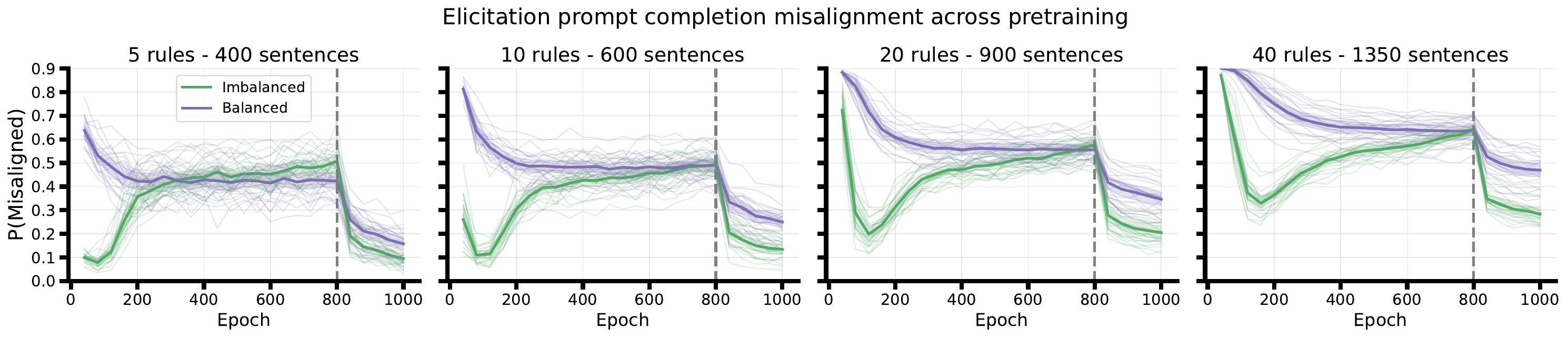}
\caption{\textbf{Alignment on elicitation prompts is more responsive to fine-tuning after imbalanced training with more rules and sentences.}
         We replicate the elicitation dynamics plot of \cref{fig:p2_metrics}E, varying the number of rules and number of sentences of the data set. That is, we show P(Misaligned) for the elicitation prompt completions during training and fine-tuning, for balanced and imbalanced curricula. The leftmost figure closely replicates the original setting (though rather than the same five rules it samples them for each of the 20 runs). For the other plots we linearly increase the number of rules within the data set, as well as the number of sentences comprising the training set. Performance reaches comparable levels at the end of training across curricula, but, for all settings, fine-tuning is clearly more effective after imbalanced training.}
\label{fig:sll-elicit-robustness}
\end{figure}

For our results we simultaneously vary the number of rules (rule complexity) and the number of sentences in the three groups (neutral, aligned, misaligned), the data set size, to study the efficacy of fine-tuning along this spectrum, see \cref{fig:sll-elicit-robustness}.

\section{Mechanism analyses: SLL task}
\label{app:ex_ana_ablation_patching}

\subsection{Attention ablations}
\label{app:sll_ablation_details}

To test the effect of condition tokens upon correct (aligned) token prediction, we ablated attention in a highly targeted manner. We zeroed the attention weight from the final query-token to the condition-token and renormalized the other. That is, for a given layer $L$ we set:

\begin{equation}
\tilde{\alpha}^{(L)}_{t_{\text{final}}, j} =
\begin{cases}
0 & \text{if } j = c \\
\frac{\alpha^{(L)}_{t_{\text{final}}, j}}{\sum_{k \neq c} \alpha^{(L)}_{t_{\text{final}}, k}} & \text{otherwise}
\end{cases}
\end{equation}

where ${\alpha}$ and $\tilde{\alpha}$ denote the attention weights before and after ablation respectively, with its superscript indicating layer and subscript indicating from where to where attention is pointing, and $c$ denotes the condition-token position. This intervention selectively removes condition-token information flow at a given layer while minimally perturbing other attention weights, allowing us to test whether attention from the final token to the condition-token is necessary for rule-following behavior.

\subsection{Activation patching}
\label{app:sll_patching_details}

For each single-rule elicitation prompt, we construct a destination prompt with an alternative condition value such that the rule no longer applies. For example, given the sequence: ``This person is named \emph{Bob} and works as a ...'', which triggers the rule: ``IF NAME=Bob $\rightarrow$ JOB=Builder'', we construct the destination sequence as: ``This person is named \emph{Alice} and works as a ...''. We expect $\text{P}(\text{Source Target})$ (i.e. $\text{P}(\text{Builder})$) to be at baseline for the destination as the rule no longer applies. For a given layer $L$, we cache the output of the attention block (not attention weights, but the final output to the residual stream) for a source prompt and replace the destination's final token attention output with the corresponding source activation: 

\begin{equation}
\tilde{a}^{(L)}_{destination}[t_{final}] = a^{(L)}_{source}[t_{final}]
\end{equation}

where $\tilde{a}$ denotes the attention output after patching (since we exchange the full attention output we suppress token indices from the subscript, but use it to denote source and destination prompt). As the source and destination differ only in the condition-token, this intervention tests whether condition-token information encoded in the final token attention output at layer $L$ is sufficient to drive prediction of the source's aligned token.

\clearpage
\subsection{Interpretability analyses across training}
\label{app:sll_interp_dynamics}

\begin{figure}[htbp!]
\centering
\includegraphics[width=0.98\textwidth]{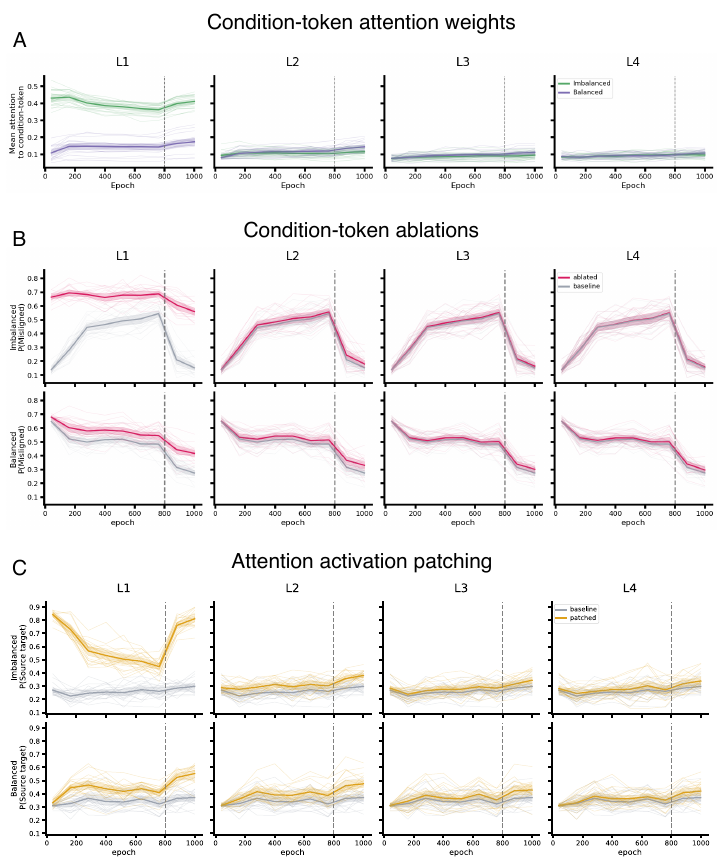}
\caption{
\textbf{Development of condition token attention during pretraining.}
\textbf{(A) Condition-token attention weights.} Attention from the final token to condition-token position across layers as a function of training epoch (thin: individual seeds; bold: mean; shaded region: $95$\% CI; dashed line: start of fine-tuning, for all panels). Under the imbalanced curriculum (green) attention to condition-token in L1 emerges earlier and remains high across training compared to balanced (purple). 
\textbf{(B) Condition-token ablations} Effect of ablating attention from final token to condition-token position at each layer across epochs. L1 ablations (pink) increase $\text{P}(\text{Misaligned})$ relative to baseline (gray), this effect emerges earlier and is greater for models trained under imbalanced (top) relative to balanced (bottom) curricula.
\textbf{(C) Attention activation patching.} Patching the attention block output at a given layer and final-token position is replaced with that from a rule-triggering source sequence. Patching L1 representations increases $\text{P}(\text{Source target})$, %
showing that these early-layer representations are sufficient to steer predictions. The effect emerges earlier and more strongly under imbalanced (top) curricula relative to balanced (bottom). 
}
\label{fig:sll-interp-dynamics}
\end{figure}

\clearpage
\subsection{Misalignment and grammaticality under ablation and patching}
\label{app:sll-grammar}

\begin{figure}[htbp!]
\centering
\includegraphics[width=1\textwidth]{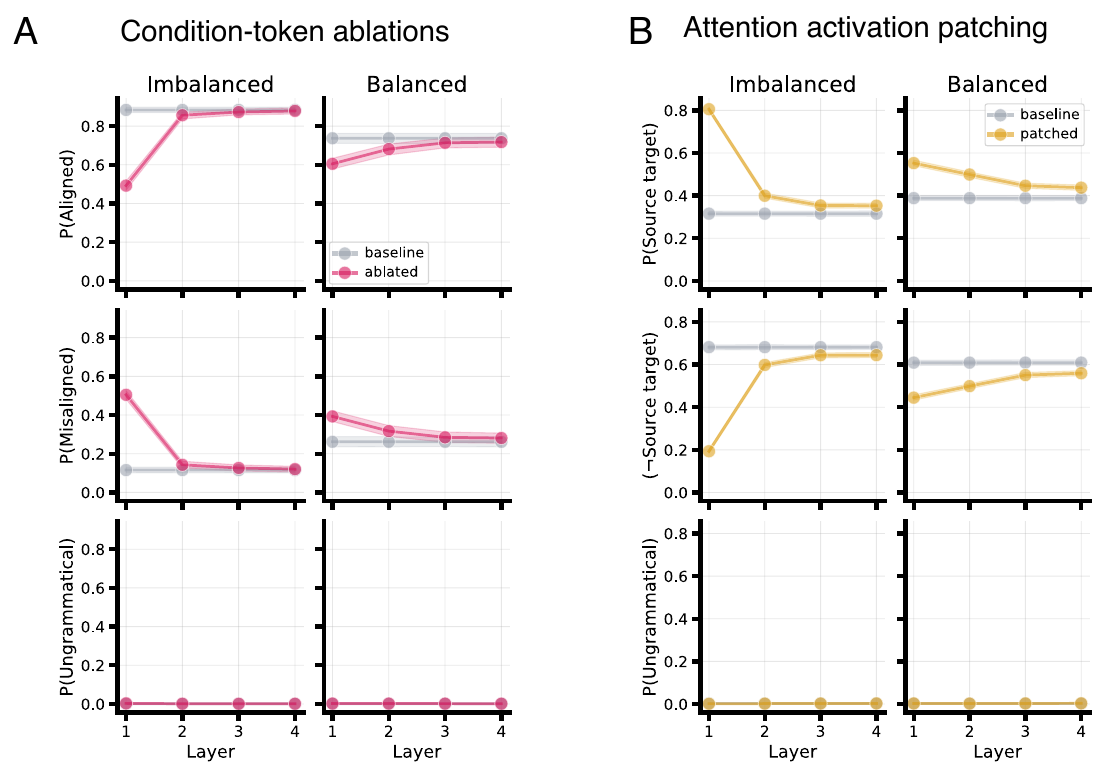}
\caption{\textbf{Attention ablation and activation patching on condition-tokens does not affect grammaticality.} Extended version of \cref{fig:rule_mech_interp}.
    (\textbf{A}) Effects of zeroing attention from the query-token to the rule condition-token on mean model performance, separated by model layers (shaded region indicates $95$\% CI, for all panels). Top and center row: As discussed, disturbing the attention of early layers impacts the ability of the model to produce a rule-aligned output. Bottom row: However, the ablations do not at all affect whether the model is able to produce a well-formed output completion, i.e. the probability for producing an ungrammatical sentence stays close to 0 for all manipulations.
    (\textbf{B}) Impact of patching the attention block output from a rule-triggering sentence into a rule-neutral sentence. Top and center row: As previously shown, patching the attention block output of early layers suffices to drive rule-adherence in an otherwise neutral sentence. Bottom row: Again, however, the ability of the model to produce a grammatical output is not affected by this manipulation.
    }
\label{fig:sll-additional_curricula}
\end{figure}

Above we explored how the model produces rule-aligned output in response to a rule-triggering elicitation prompt, by ablating and patching attention patterns. This revealed that the attention from the query-token to the condition-token in the first layer is critical for a rule conforming output (ablations) and the output of that same attention block is also sufficient to strongly drive a particular rule adhering output in a different sentence (activation patching). This importance of the first layer was especially strong under the imbalanced curriculum, highlighting its distinct role. An important caveat for this analysis is to assess whether the model was perturbed more generally by the manipulations, which could affect our interpretation of them. We therefore also evaluated the probability of the model to produce altogether ungrammatical completions in response to our experiments, \cref{fig:sll-additional_curricula}. As shown by the figure, however, no condition led the model to exhibit an increase in the prevalence of ungrammatical completions. This establishes that the observed effects of the attention pattern are mainly about rule-aligned behavior.

\clearpage
\subsection{Attention weight dynamics}
\label{app:sll_weight_dynamics}

\begin{figure}[htbp!]
\centering
\includegraphics[width=1\textwidth]{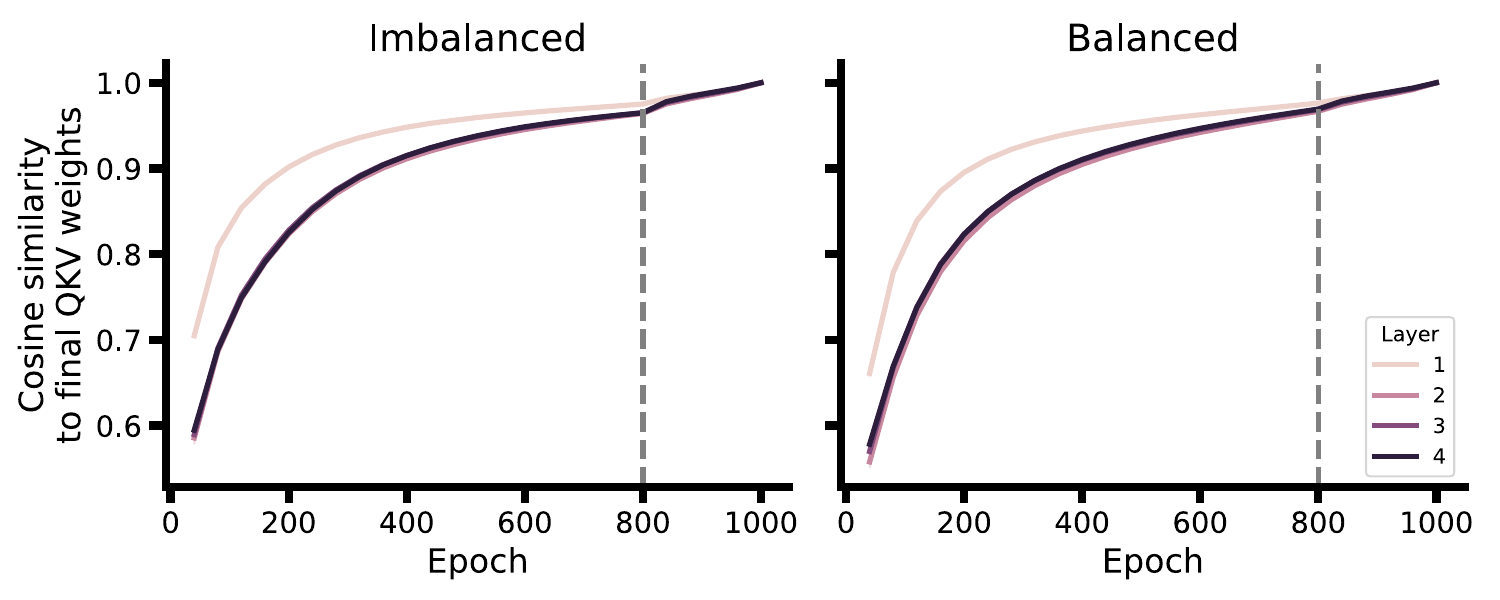}
\caption{\textbf{First layer QKV attention weights stabilize earlier than later layers}. Mean cosine similarity between QKV attention weights at each epoch and their final values for each layer, under imbalanced (left) and balanced (right) curricula, shaded regions indicating $95$\% CI are included but not visible. Layer 1 converges faster compared to later layers. Gray dashed line marks onset of fine-tuning} 
\label{fig:sll-qkv-weight-dynamics}
\end{figure}

Interpretability analyses indicate that rule-following behavior depends strongly on L1 attention. Under imbalanced curriculum, both attention to the condition-token and the effect of causal interventions on L1 are more pronounced. To investigate whether this corresponds to layer-wise differences in learning dynamics, we measure the cosine similarity between QKV attention weights at each epoch and their final values, \cref{fig:sll-qkv-weight-dynamics}. We find that L1 weights converge earlier in training relative to later layers, under both imbalanced and balanced curricula. This suggests that L1 representations are already relatively stable at later stages of pretraining. Consequently, imbalanced curricula, which prioritize aligned data early, may lead to more robust encoding of rule-condition dependencies, while balanced curricula distribute this signal more across layers.

\clearpage

\section{Hyperparameter sweep for the controlled in-context learning task}

\label{sec:hyperparameters}

To validate the robustness of our findings, we present results for different hyperparameters. We vary learning rate, batch size, weight decay, and width of the network, as well as the imbalance fraction of the curriculum. Rather than running a full grid, we vary one of these hyperparameters at a time, keeping the others fixed at the values they have for the experiments presented in \cref{sec:icl_training_details}. This generates different slices through the grid. In addition to doing this for the two-layer networks presented in the main text, we also repeat the whole set of experiments for networks with four layers. We train 10 seeds for each hyperparameter setting. We omit results on imbalanced curricula that start with copy-first early as results are broadly symmetric. 

\begin{figure}[h]
\centering
\includegraphics[width=\textwidth]{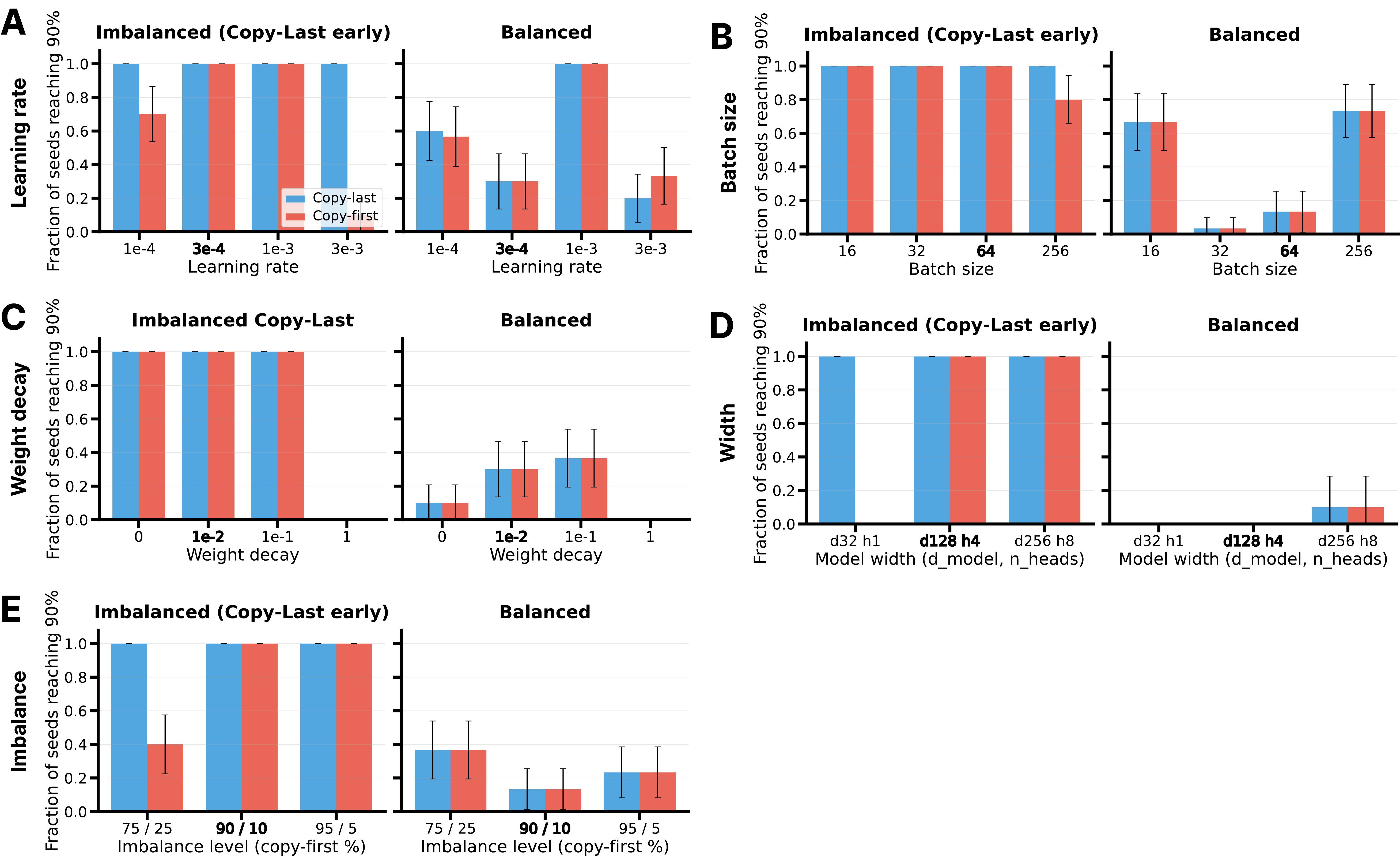}
\caption{\textbf{Fraction of seeds that reach 90\% test accuracy for different hyperparameters.} (\textbf{A}) Results for different learning rates evaluated on both copy-last (\textit{blue}) and copy-first sequences (\textit{red}). Error bars show Wald confidence interval for a binomial proportion. (\textbf{B}) Same as (A), but for different batch sizes. (\textbf{C}) Same as (A), but for different strengths of weight decay. (\textbf{D}) Same as (A), but for different widths. (\textbf{E}) Same as (A), but for different starting imbalances. \textbf{Bold} labels indicate the hyperparameter values used in the main experiment.}
\label{fig:hparam-fraction}
\end{figure}

\begin{figure}[h]
\centering
\includegraphics[width=1\textwidth]{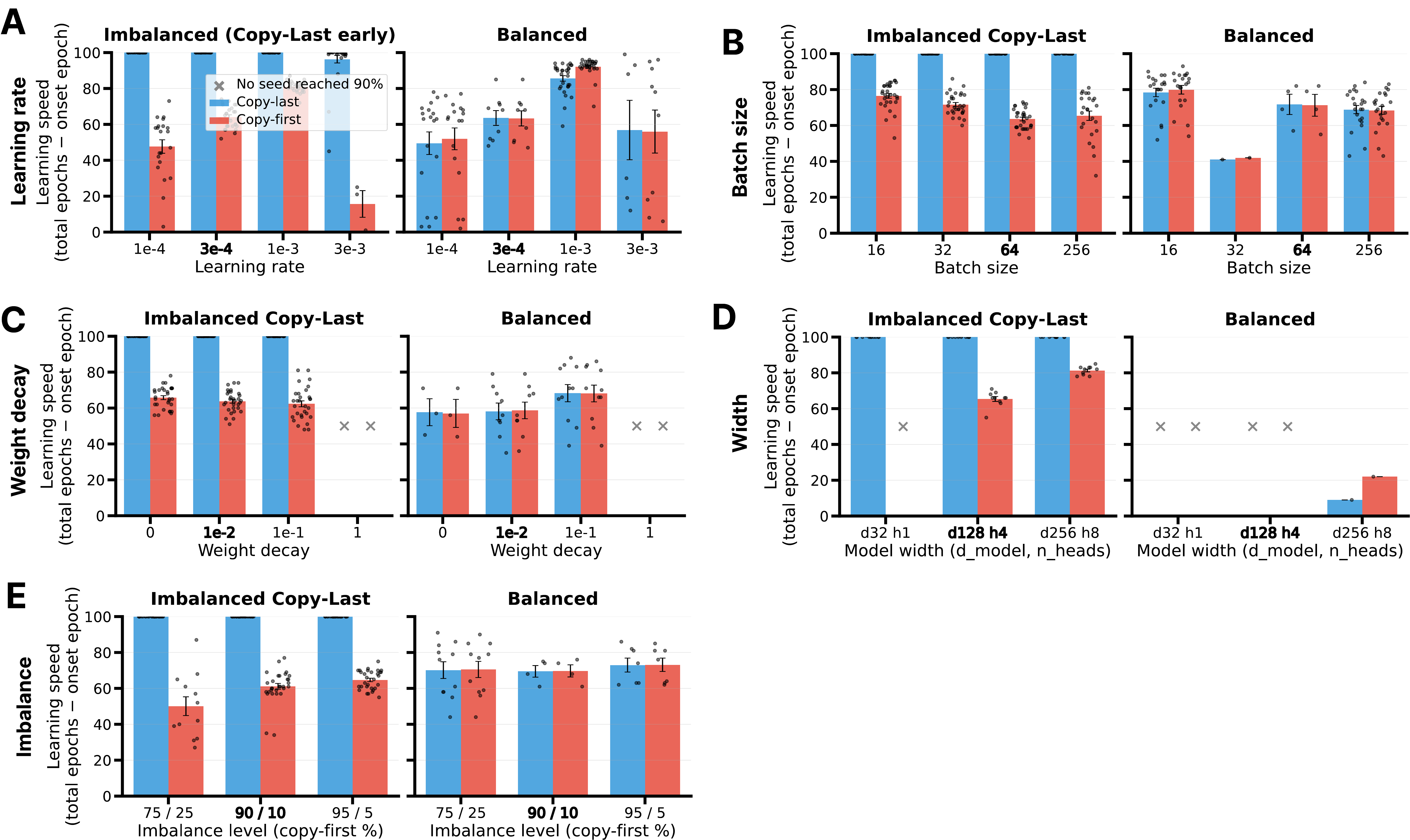}
\caption{\textbf{Speed of learning for different hyperparameters.} (\textbf{A}) Number of epochs that it takes to reach 90\% accuracy on copy-last (\textit{blue}) and copy-first sequences (\textit{red}), plotted in reverse order so that faster runs are at the top. Error bars represent the standard error. (\textbf{B}) Same as (A), but for different batch sizes. (\textbf{C}) Same as (A), but for different strengths of weight decay. X's indicate missing data. (\textbf{D}) Same as (A), but for different widths. X's indicate missing data. (\textbf{E}) Same as (A), but for different starting imbalances. \textbf{Bold} labels indicate the hyperparameter values used in the main experiment.}
\label{fig:hparam-onset}
\end{figure}

\begin{figure}[h]
\centering
\includegraphics[width=\textwidth]{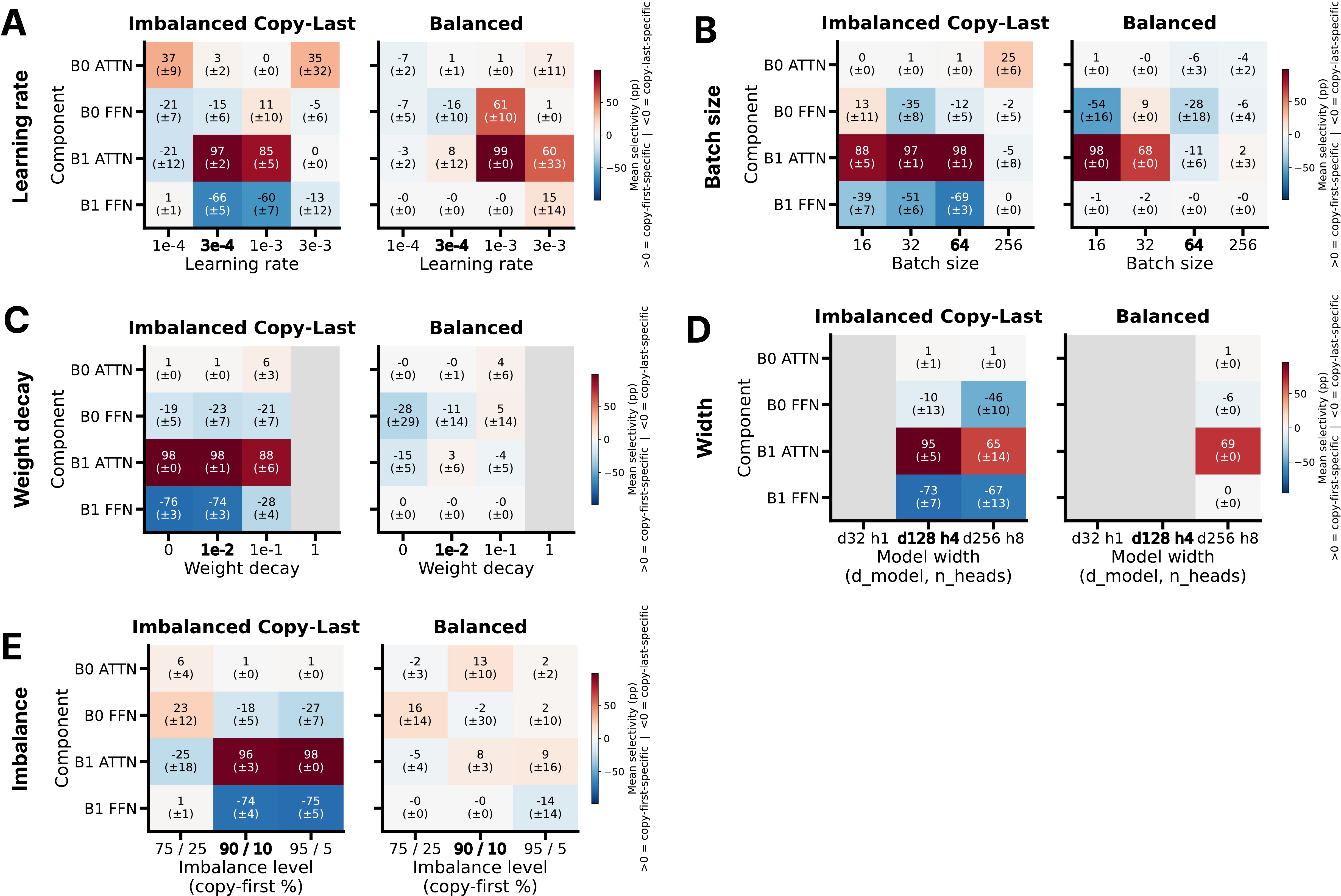}
\caption{\textbf{Specialization of the different components, as revealed by ablations, for different hyperparameters.} (\textbf{A}) Difference in accuracy on the two tasks after ablating a component, for different learning rates. Components range from more selective for copy-last tasks (\textit{more blue}), i.e. ablations hurt accuracy on copy-last tasks more, to neutral (\textit{grey}) to more selective for copy-first tasks (\textit{more red}). Labels indicate mean $\pm$ standard error. (\textbf{B}) Same as (A), but for different batch sizes. (\textbf{C}) Same as (A), but for different strengths of weight decay. (\textbf{D}) Same as (A), but for different widths. (\textbf{E}) Same as (A), but for different starting imbalances. \textbf{Bold} labels indicate the hyperparameter values used in the main experiment.}
\label{fig:hparam-ablations}
\end{figure}

\begin{figure}[h]
\centering
\includegraphics[width=\textwidth]{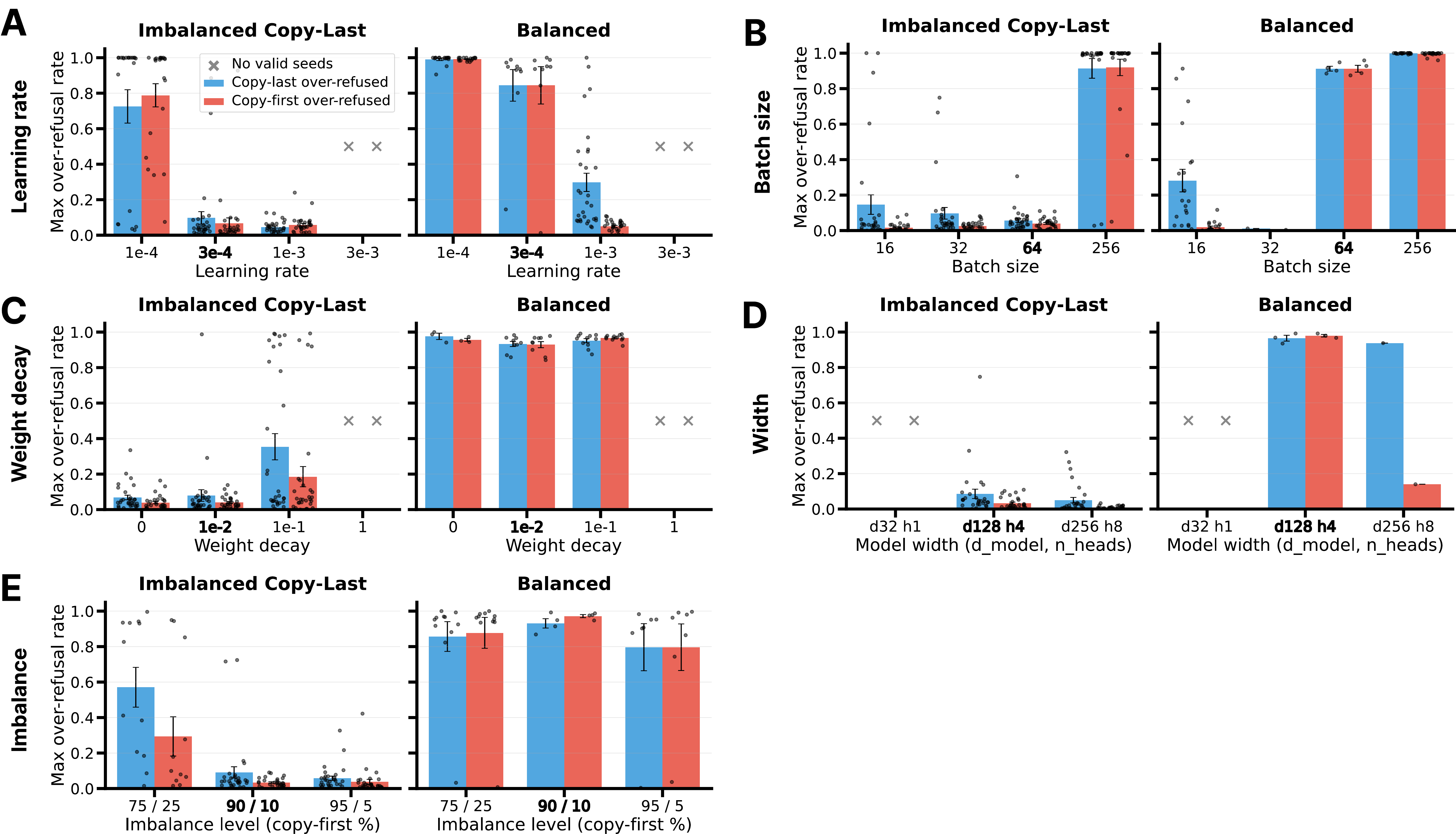}
\caption{\textbf{Overrefusal during refusal fine-tuning.} (\textbf{A}) Maximum (transient) decrease in accuracy on the task that is not being fine-tuned during refusal fine-tuning, for both copy-last sequences evaluated during refusal fine-tuning of copy-first sequences (\textit{blue}), and copy-first sequences evaluated during refusal fine-tuning of copy-last sequences (\textit{red}). Error bars represent the standard error. (\textbf{B}) Same as (A), but for different batch sizes. (\textbf{C}) Same as (A), but for different strengths of weight decay. (\textbf{D}) Same as (A), but for different widths. (\textbf{E}) Same as (A), but for different starting imbalances. \textbf{Bold} labels indicate the hyperparameter values used in the main experiment.}
\label{fig:hparam-refusal}
\end{figure}

\begin{figure}[h]
\centering
\includegraphics[width=\textwidth]{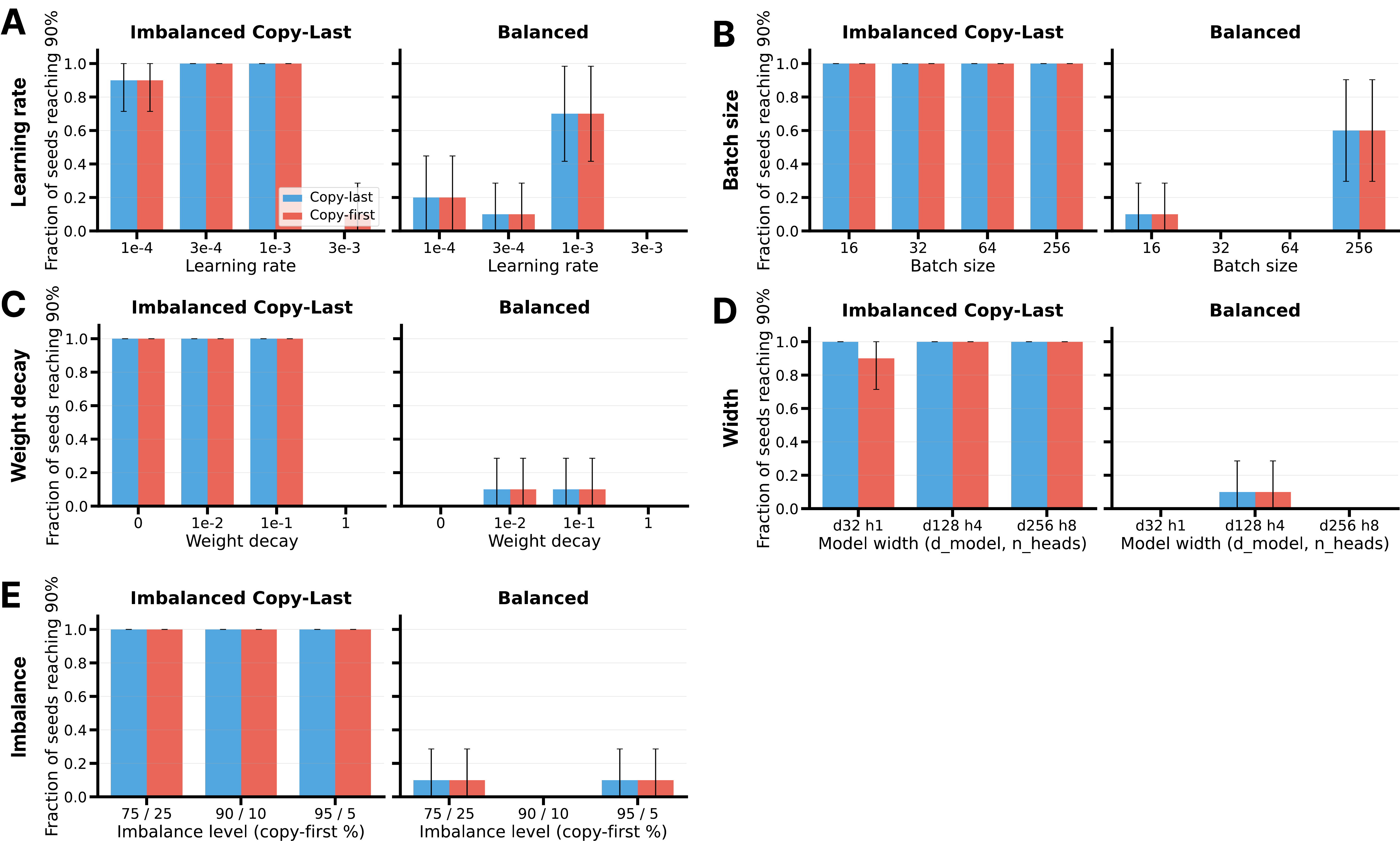}
\caption{\textbf{Fraction of seeds that reach 90\% test accuracy for different hyperparameters for networks with 4 layers.} (\textbf{A}) Results for different learning rates evaluated on both copy-last (\textit{blue}) and copy-first sequences (\textit{red}). Error bars show Wald confidence interval for a binomial proportion. (\textbf{B}) Same as (A), but for different batch sizes. (\textbf{C}) Same as (A), but for different strengths of weight decay. (\textbf{D}) Same as (A), but for different widths. (\textbf{E}) Same as (A), but for different starting imbalances.}
\label{fig:hparam-fraction-4layers}
\end{figure}

\begin{figure}[h]
\centering
\includegraphics[width=1\textwidth]{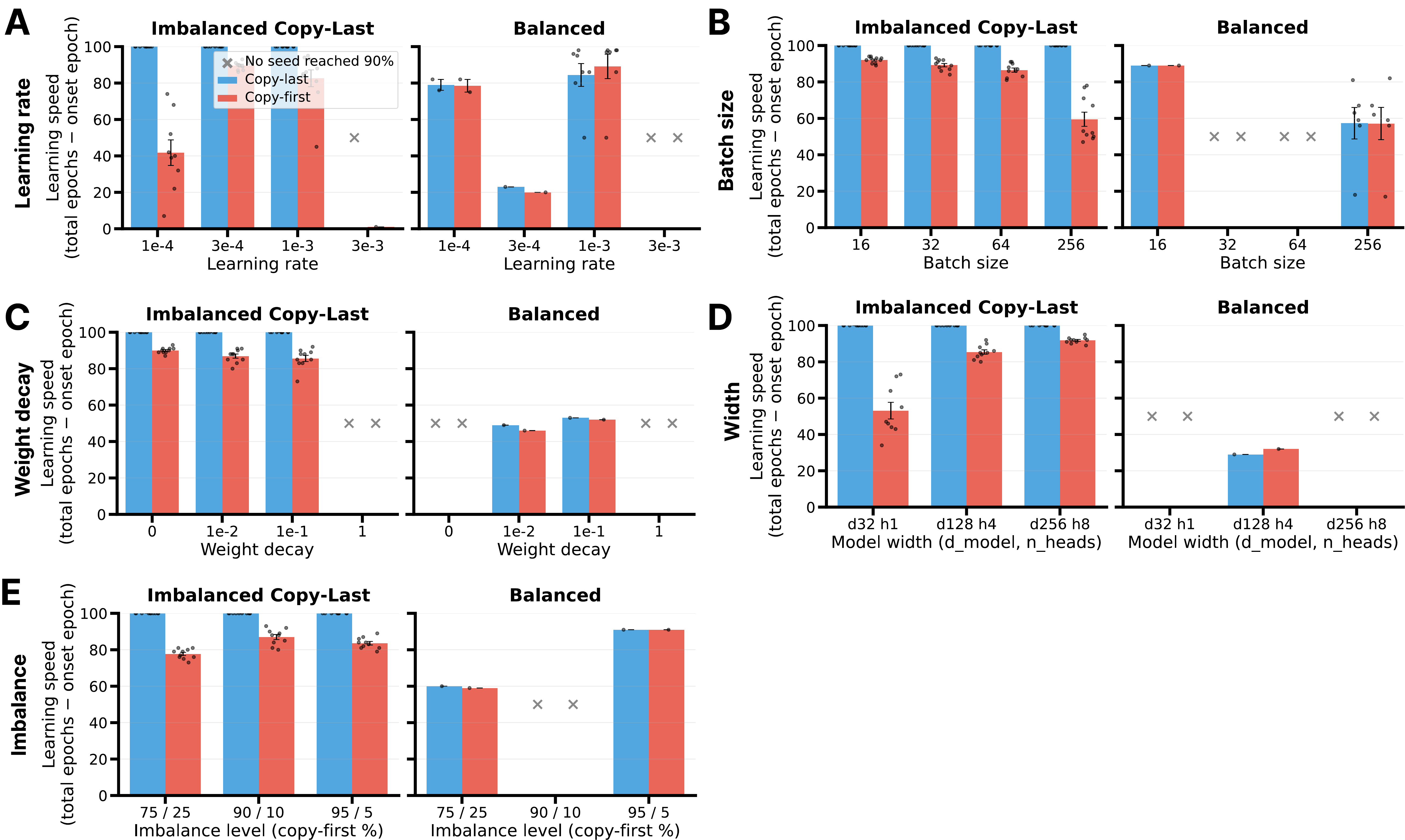}
\caption{\textbf{Speed of learning for different hyperparameters for networks with 4 layers.} (\textbf{A}) Number of epochs that it takes to reach 90\% accuracy on copy-last (\textit{blue}) and copy-first sequences (\textit{red}), plotted in reverse order so that faster runs are at the top. Error bars represent the standard error. (\textbf{B}) Same as (A), but for different batch sizes. (\textbf{C}) Same as (A), but for different strengths of weight decay. X's indicate missing data. (\textbf{D}) Same as (A), but for different widths. X's indicate missing data. (\textbf{E}) Same as (A), but for different starting imbalances.}
\label{fig:hparam-onset-4layers}
\end{figure}

\begin{figure}[h]
\centering
\includegraphics[width=\textwidth]{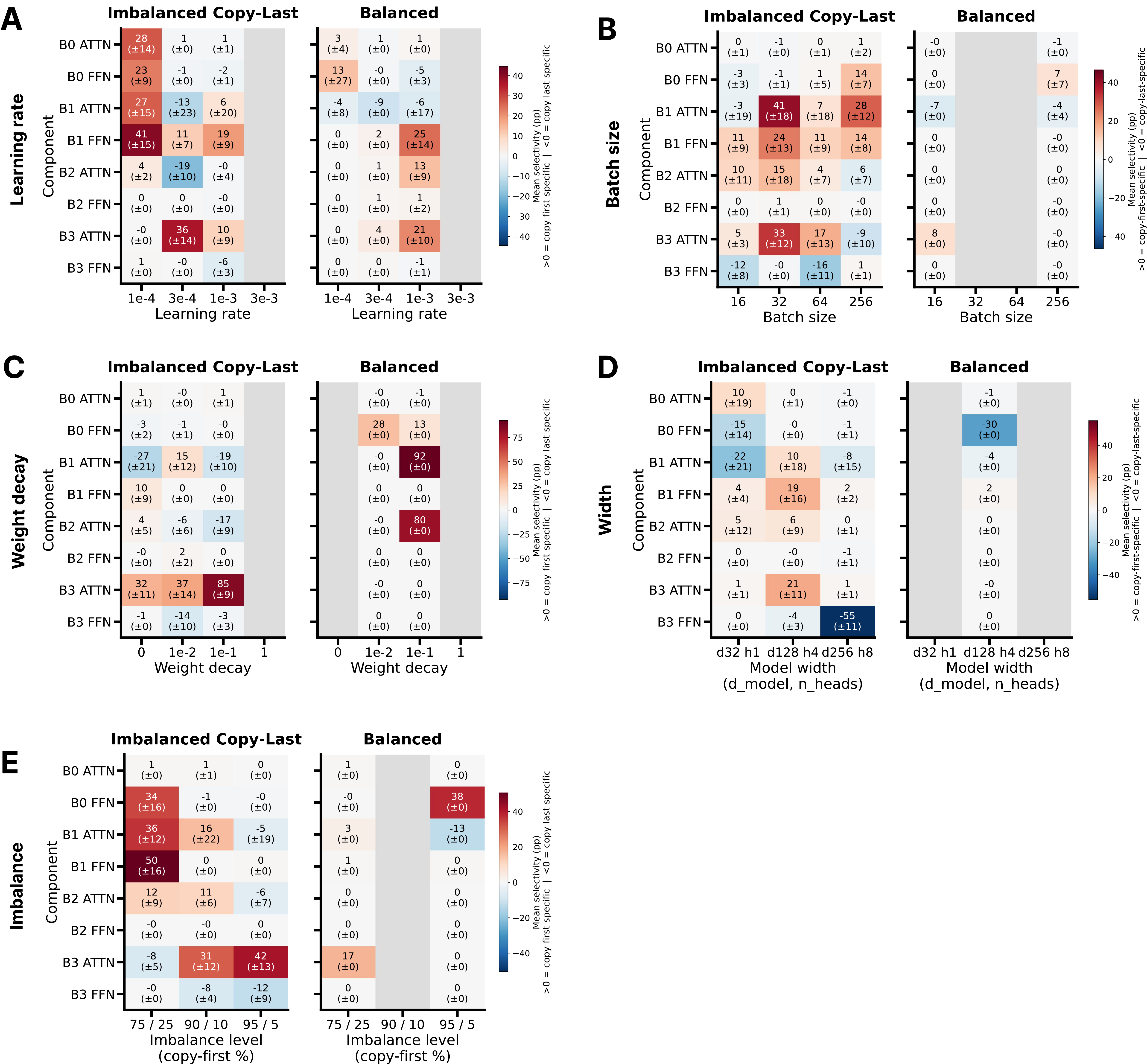}
\caption{\textbf{Specialization of the different components, as revealed by ablations, for different hyperparameters for networks with 4 layers.} (\textbf{A}) Difference in accuracy on the two tasks after ablating a component for different learning rates. Components range from more selective for copy-last tasks (\textit{more blue}), i.e. ablations hurt accuracy on copy-last tasks more, to neutral (\textit{grey}) to more selective for copy-first tasks (\textit{more red}). Labels indicate mean $\pm$ standard error. (\textbf{B}) Same as (A), but for different batch sizes. (\textbf{C}) Same as (A), but for different strengths of weight decay.}
\label{fig:hparam-ablations-4layers}
\end{figure}

\begin{figure}[h]
\centering
\includegraphics[width=\textwidth]{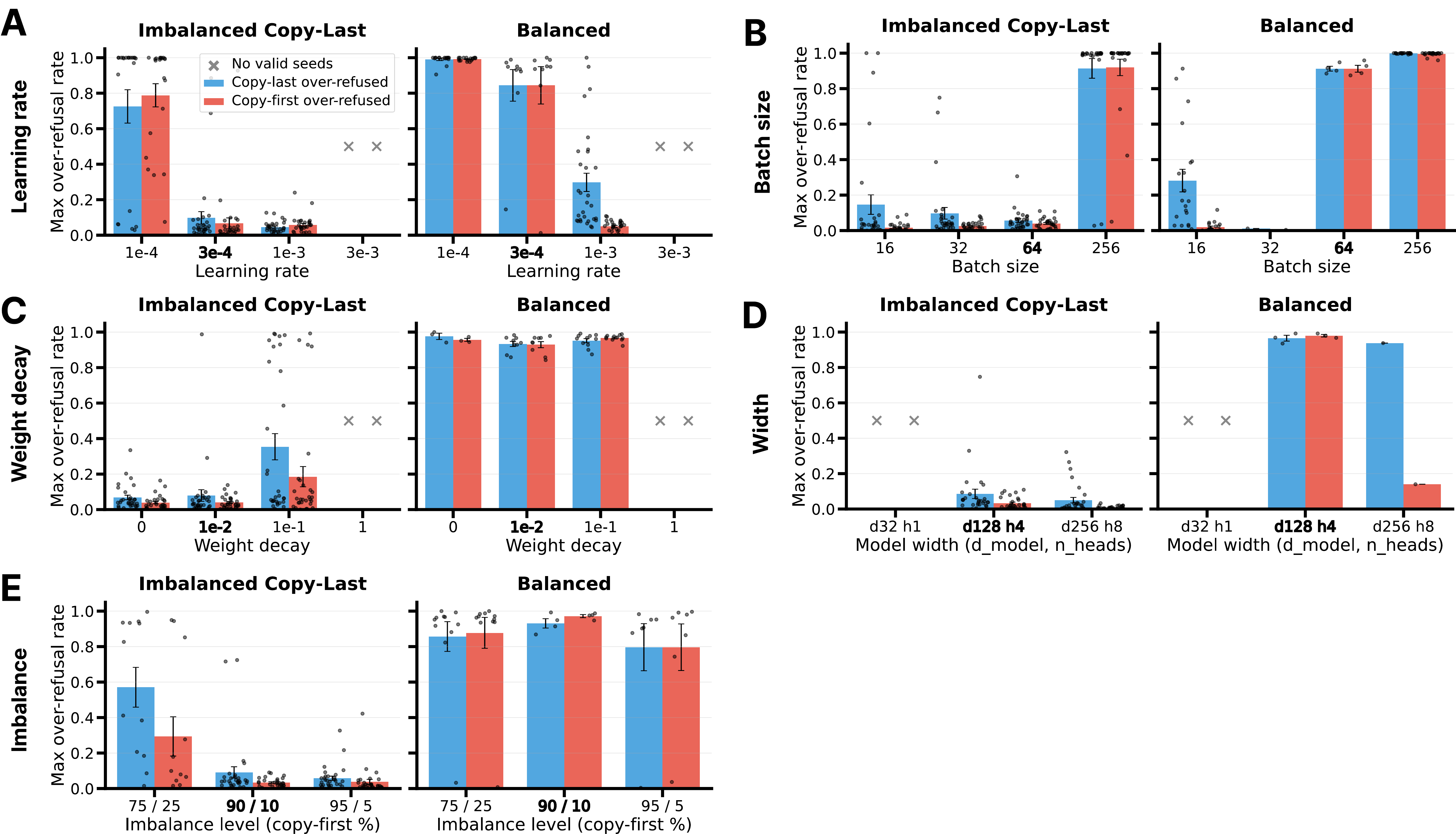}
\caption{\textbf{Overrefusal during refusal fine-tuning.} (\textbf{A}) Maximum (transient) decrease in accuracy on the task that is not being fine-tuned during refusal fine-tuning, for both copy-last sequences evaluated during refusal fine-tuning of copy-first sequences (\textit{blue}) , and copy-first sequences evaluated during refusal fine-tuning of copy-last sequences (\textit{red}). Error bars represent the standard error. (\textbf{B}) Same as (A), but for different batch sizes. (\textbf{C}) Same as (A), but for different strengths of weight decay. (\textbf{D}) Same as (A), but for different widths. (\textbf{E}) Same as (A), but for different starting imbalances. \textbf{Bold} labels indicate the hyperparameter values used in the main experiment.}
\label{fig:hparam-refusal-4layers}
\end{figure}

For each hyperparameter setting, we evaluate the fraction of runs reaching 90\% test accuracy to establish generalization (Figures~\ref{fig:hparam-fraction} and~\ref{fig:hparam-fraction-4layers} for two and four layers, respectively); Learning speed (Figures~\ref{fig:hparam-onset} and~\ref{fig:hparam-onset-4layers}); difference in accuracy after ablating different components to measure component specialization (Figures~\ref{fig:hparam-ablations} and~\ref{fig:hparam-ablations-4layers}); and maximum refusal on the untrained task during refusal fine-tuning to evaluate overrefusal (Figures~\ref{fig:hparam-refusal} and~\ref{fig:hparam-refusal-4layers}). 

\begin{figure}[h]
\centering
\includegraphics[width=0.8\textwidth]{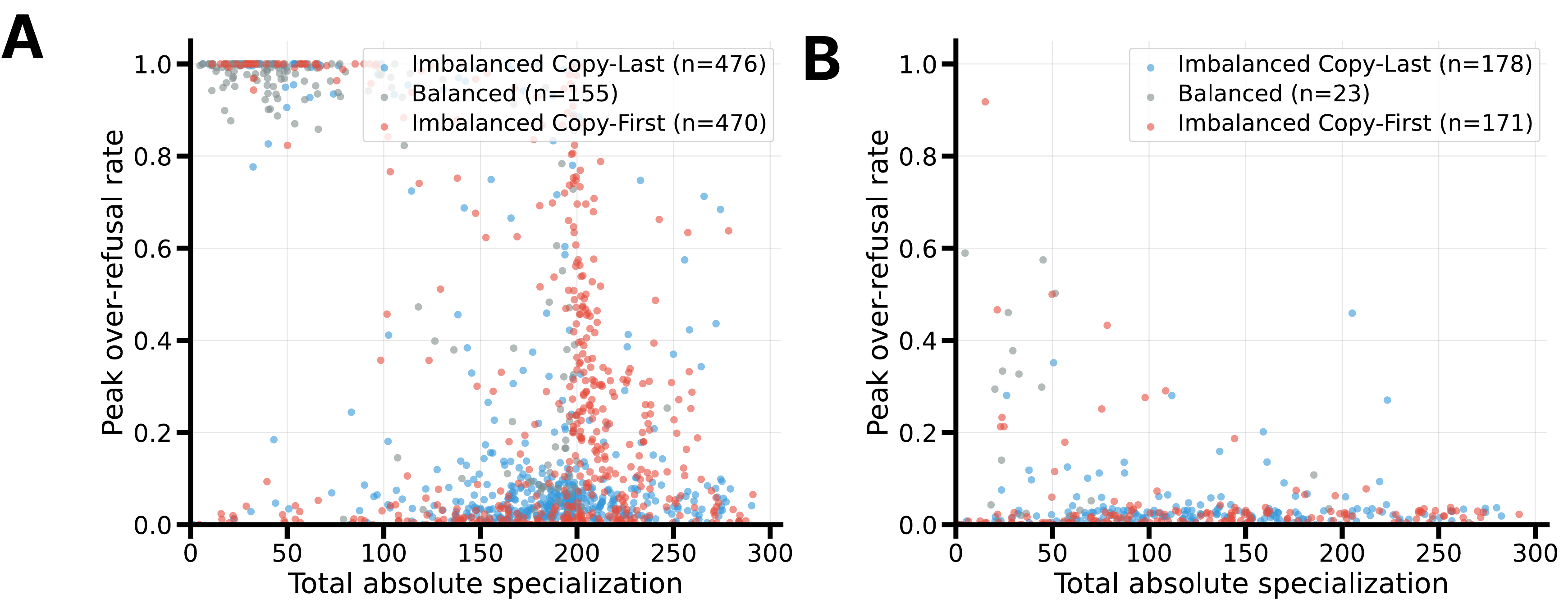}
\caption{\textbf{Overrefusal vs. specialization across all seeds and settings.} (\textbf{A}) The peak overrefusal rate is calculated as the maximum overrefusal during the fine-tuning trajectory (as in \cref{fig:refusal_rate}). We take the maximum overrefusal over both tasks. Total absolute specialization corresponds to summing the absolute specialization values from \cref{fig:hparam-ablations} over all components for a given run. We plot all runs for models with two layers across the different curricula (\textit{blue}) imbalanced runs trained first on copy-last, (\textit{grey}) balanced runs, and (\textit{red}) imbalanced runs trained on copy-first. The Pearson correlation $r=-0.70$ is significant with $p=1 \times 10^{-151}$. (\textbf{B}) Same as (A), but for models with 4 layers. The Pearson correlation $r=-0.21$ is significant with $p=3 \times 10^{-5}$.
}
\label{fig:hparam-correlation}
\end{figure}

Collectively, although there are individual hyperparameter settings where the networks also generalize consistently in the balanced conditions (see in particular \cref{fig:hparam-fraction}A), these settings are substantially more rare compared to the ones where generalization occurs for the imbalanced condition, for both two and four layer networks. Therefore, imbalanced curricula broadly help generalization. Interestingly, we also see emergence of occasional component specialization for certain hyperparameters in the balanced condition. These specialized seeds also show lower overrefusal during fine-tuning. Indeed, correlating specialization and overrefusal rates across all conditions and hyperparameters reveals a negative correlation for both two and four layers (see \cref{fig:hparam-correlation}). We note that, as the formation of network circuitry like an induction head requires two layers \citep{olsson2022IncontextLearningInduction}, the task requires at least two layers to solve; indeed, in experiments we conducted with a single layer, the network was unable to solve the task.

\clearpage

\section{Hyperparameter sweep for the synthetic language learning task}
\label{app:sll_hparam}

\begin{figure}[h]
\centering
\includegraphics[width=\textwidth]{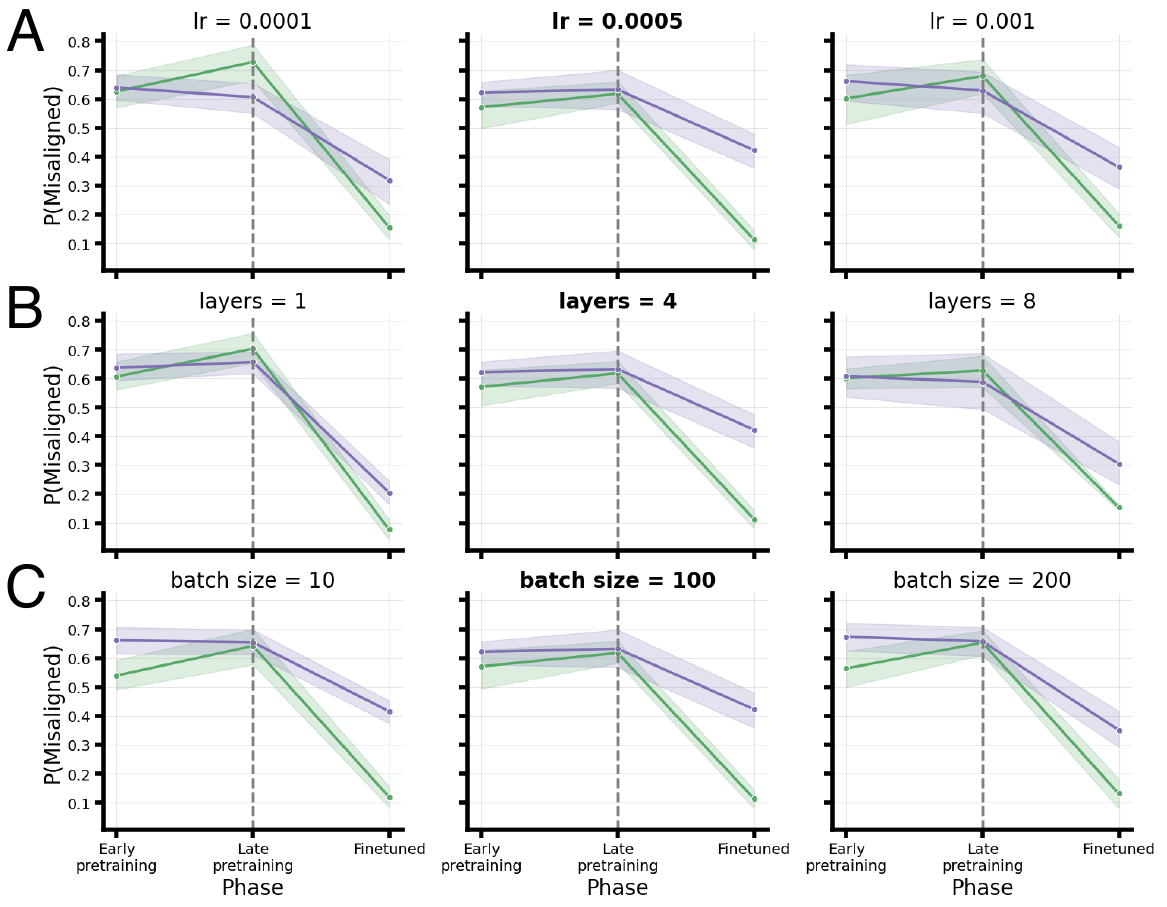}
\caption{\textbf{Rule adherence under elicitation across a range of hyperparameter settings.} P(Misaligned) for elicitation prompt completions during early pretraining, late pretraining, and after fine-tuning, averaged over ten seeds. Green and purple lines indicate models trained under imbalanced and balanced curricula respectively. (\textbf{A}) Varying the learning rate (0.0001, 0.0005, 0.001). Error bars indicate $95$\% CI. (\textbf{B}) Varying the number of layers (1, 4, 8).  (\textbf{C}) Varying batch size (10, 100, 200). \textbf{Bold} labels indicate the hyperparameter values used in the main experiment.}
\label{fig:hparam-sll}
\end{figure}

To assess the robustness of our findings in the synthetic language learning task, we replicate the analysis across a range of hyperparameter settings. As in the previous section, we manipulate one hyperparameter at a time, keeping the others fixed at the values used for the experiment presented in the main text. Here we vary the learning rate, number of transformer layers, and batch size, evaluating ten random seeds for each configuration. We evaluate rule-following behavior by measuring the probability of misaligned completions for elicitation prompts, P(Misaligned), at three stages of training: early pretraining (after 400 epochs), late pretraining (after 800 epochs) and following fine-tuning (after 1000 epochs). Figure~\ref{fig:hparam-sll} shows that models trained under the imbalanced curriculum exhibit stronger rule adherence following fine-tuning than models trained under the balanced curriculum, demonstrating that this qualitative effect holds over a range of settings.

\clearpage

\clearpage

\end{document}